\documentclass[]{thisstyle_purple}

\usepackage{amsmath,amssymb}
\usepackage{multirow}
\usepackage{bigdelim}
\usepackage{todonotes}
\usepackage{longtable}
\usepackage{tabularray}
\usepackage{wrapfig}
\usepackage{fontawesome5}
\usepackage[most]{tcolorbox}
\tcbuselibrary{listings, skins, breakable}
\usepackage[dvipsnames,svgnames,table]{xcolor}
\usepackage{url}
\usepackage{xspace}
\usepackage{fancyvrb}
\usepackage{fvextra}
\usepackage{caption}
\usepackage{graphicx}
\usepackage{svg}
\usepackage{hyperref}
\usepackage{soul}
\usepackage{enumitem}
\usepackage{tgpagella}
\usepackage{color-edits}

\addauthor{ss}{oc-orange-9}

\definecolor{oc-gray-0}{HTML}{F8F9FA}
\definecolor{oc-gray-1}{HTML}{F1F3F5}
\definecolor{oc-gray-2}{HTML}{E9ECEF}
\definecolor{oc-gray-3}{HTML}{DEE2E6}
\definecolor{oc-gray-4}{HTML}{CED4DA}
\definecolor{oc-gray-5}{HTML}{ADB5BD}
\definecolor{oc-gray-6}{HTML}{868E96}
\definecolor{oc-gray-7}{HTML}{495057}
\definecolor{oc-gray-8}{HTML}{343A40}
\definecolor{oc-gray-9}{HTML}{212529}

\definecolor{oc-red-0}{HTML}{FFF5F5}
\definecolor{oc-red-1}{HTML}{FFE3E3}
\definecolor{oc-red-2}{HTML}{FFC9C9}
\definecolor{oc-red-3}{HTML}{FFA8A8}
\definecolor{oc-red-4}{HTML}{FF8787}
\definecolor{oc-red-5}{HTML}{FF6B6B}
\definecolor{oc-red-6}{HTML}{FA5252}
\definecolor{oc-red-7}{HTML}{F03E3E}
\definecolor{oc-red-8}{HTML}{E03131}
\definecolor{oc-red-9}{HTML}{C92A2A}

\definecolor{oc-pink-0}{HTML}{FFF0F6}
\definecolor{oc-pink-1}{HTML}{FFDEEB}
\definecolor{oc-pink-2}{HTML}{FCC2D7}
\definecolor{oc-pink-3}{HTML}{FAA2C1}
\definecolor{oc-pink-4}{HTML}{F783AC}
\definecolor{oc-pink-5}{HTML}{F06595}
\definecolor{oc-pink-6}{HTML}{E64980}
\definecolor{oc-pink-7}{HTML}{D6336C}
\definecolor{oc-pink-8}{HTML}{C2255C}
\definecolor{oc-pink-9}{HTML}{A61E4D}

\definecolor{oc-grape-0}{HTML}{F8F0FC}
\definecolor{oc-grape-1}{HTML}{F3D9FA}
\definecolor{oc-grape-2}{HTML}{EEBEFA}
\definecolor{oc-grape-3}{HTML}{E599F7}
\definecolor{oc-grape-4}{HTML}{DA77F2}
\definecolor{oc-grape-5}{HTML}{CC5DE8}
\definecolor{oc-grape-6}{HTML}{BE4BDB}
\definecolor{oc-grape-7}{HTML}{AE3EC9}
\definecolor{oc-grape-8}{HTML}{9C36B5}
\definecolor{oc-grape-9}{HTML}{862E9C}

\definecolor{oc-violet-0}{HTML}{F3F0FF}
\definecolor{oc-violet-1}{HTML}{E5DBFF}
\definecolor{oc-violet-2}{HTML}{D0BFFF}
\definecolor{oc-violet-3}{HTML}{B197FC}
\definecolor{oc-violet-4}{HTML}{9775FA}
\definecolor{oc-violet-5}{HTML}{845EF7}
\definecolor{oc-violet-6}{HTML}{7950F2}
\definecolor{oc-violet-7}{HTML}{7048E8}
\definecolor{oc-violet-8}{HTML}{6741D9}
\definecolor{oc-violet-9}{HTML}{5F3DC4}

\definecolor{oc-indigo-0}{HTML}{EDF2FF}
\definecolor{oc-indigo-1}{HTML}{DBE4FF}
\definecolor{oc-indigo-2}{HTML}{BAC8FF}
\definecolor{oc-indigo-3}{HTML}{91A7FF}
\definecolor{oc-indigo-4}{HTML}{748FFC}
\definecolor{oc-indigo-5}{HTML}{5C7CFA}
\definecolor{oc-indigo-6}{HTML}{4C6EF5}
\definecolor{oc-indigo-7}{HTML}{4263EB}
\definecolor{oc-indigo-8}{HTML}{3B5BDB}
\definecolor{oc-indigo-9}{HTML}{364FC7}

\definecolor{oc-blue-0}{HTML}{E7F5FF}
\definecolor{oc-blue-1}{HTML}{D0EBFF}
\definecolor{oc-blue-2}{HTML}{A5D8FF}
\definecolor{oc-blue-3}{HTML}{74C0FC}
\definecolor{oc-blue-4}{HTML}{4DABF7}
\definecolor{oc-blue-5}{HTML}{339AF0}
\definecolor{oc-blue-6}{HTML}{228BE6}
\definecolor{oc-blue-7}{HTML}{1C7ED6}
\definecolor{oc-blue-8}{HTML}{1971C2}
\definecolor{oc-blue-9}{HTML}{1864AB}

\definecolor{oc-cyan-0}{HTML}{E3FAFC}
\definecolor{oc-cyan-1}{HTML}{C5F6FA}
\definecolor{oc-cyan-2}{HTML}{99E9F2}
\definecolor{oc-cyan-3}{HTML}{66D9E8}
\definecolor{oc-cyan-4}{HTML}{3BC9DB}
\definecolor{oc-cyan-5}{HTML}{22B8CF}
\definecolor{oc-cyan-6}{HTML}{15AABF}
\definecolor{oc-cyan-7}{HTML}{1098AD}
\definecolor{oc-cyan-8}{HTML}{0C8599}
\definecolor{oc-cyan-9}{HTML}{0B7285}

\definecolor{oc-teal-0}{HTML}{E6FCF5}
\definecolor{oc-teal-1}{HTML}{C3FAE8}
\definecolor{oc-teal-2}{HTML}{96F2D7}
\definecolor{oc-teal-3}{HTML}{63E6BE}
\definecolor{oc-teal-4}{HTML}{38D9A9}
\definecolor{oc-teal-5}{HTML}{20C997}
\definecolor{oc-teal-6}{HTML}{12B886}
\definecolor{oc-teal-7}{HTML}{0CA678}
\definecolor{oc-teal-8}{HTML}{099268}
\definecolor{oc-teal-9}{HTML}{087F5B}

\definecolor{oc-green-0}{HTML}{EBFBEE}
\definecolor{oc-green-1}{HTML}{D3F9D8}
\definecolor{oc-green-2}{HTML}{B2F2BB}
\definecolor{oc-green-3}{HTML}{8CE99A}
\definecolor{oc-green-4}{HTML}{69DB7C}
\definecolor{oc-green-5}{HTML}{51CF66}
\definecolor{oc-green-6}{HTML}{40C057}
\definecolor{oc-green-7}{HTML}{37B24D}
\definecolor{oc-green-8}{HTML}{2F9E44}
\definecolor{oc-green-9}{HTML}{2B8A3E}

\definecolor{oc-lime-0}{HTML}{F4FCE3}
\definecolor{oc-lime-1}{HTML}{E9FAC8}
\definecolor{oc-lime-2}{HTML}{D8F5A2}
\definecolor{oc-lime-3}{HTML}{C0EB75}
\definecolor{oc-lime-4}{HTML}{A9E34B}
\definecolor{oc-lime-5}{HTML}{94D82D}
\definecolor{oc-lime-6}{HTML}{82C91E}
\definecolor{oc-lime-7}{HTML}{74B816}
\definecolor{oc-lime-8}{HTML}{66A80F}
\definecolor{oc-lime-9}{HTML}{5C940D}

\definecolor{oc-yellow-0}{HTML}{FFF9DB}
\definecolor{oc-yellow-1}{HTML}{FFF3BF}
\definecolor{oc-yellow-2}{HTML}{FFEC99}
\definecolor{oc-yellow-3}{HTML}{FFE066}
\definecolor{oc-yellow-4}{HTML}{FFD43B}
\definecolor{oc-yellow-5}{HTML}{FCC419}
\definecolor{oc-yellow-6}{HTML}{FAB005}
\definecolor{oc-yellow-7}{HTML}{F59F00}
\definecolor{oc-yellow-8}{HTML}{F08C00}
\definecolor{oc-yellow-9}{HTML}{E67700}

\definecolor{oc-orange-0}{HTML}{FFF4E6}
\definecolor{oc-orange-1}{HTML}{FFE8CC}
\definecolor{oc-orange-2}{HTML}{FFD8A8}
\definecolor{oc-orange-3}{HTML}{FFC078}
\definecolor{oc-orange-4}{HTML}{FFA94D}
\definecolor{oc-orange-5}{HTML}{FF922B}
\definecolor{oc-orange-6}{HTML}{FD7E14}
\definecolor{oc-orange-7}{HTML}{F76707}
\definecolor{oc-orange-8}{HTML}{E8590C}
\definecolor{oc-orange-9}{HTML}{D9480F}

\newlength{\myparskiplength}
\newlength{\myparindentlength}
\parindent = \myparindentlength
\usepackage{booktabs}
\usepackage{enumitem}
\usepackage{algorithm}
\usepackage{algpseudocode}

\definecolor{purple}{HTML}{8B51A8}
\definecolor{darkblue}{RGB}{33, 95, 154}
\definecolor{myblue}{RGB}{61, 124, 209}
\definecolor{darkpurple}{RGB}{115, 109, 160}
\definecolor{myyellow}{RGB}{217, 145, 45}
\definecolor{mygreen}{RGB}{110, 172, 29}
\definecolor{mypink}{RGB}{194, 35, 93}
\definecolor{mygrey}{HTML}{868D96}

\definecolor{retained}{HTML}{215F9A}
\definecolor{abandoned}{HTML}{C2235D}
\definecolor{superseded}{HTML}{E17F00}
\definecolor{pinkpurple}{HTML}{7030A0}
\definecolor{lightpurple}{HTML}{ECE2FD}

\DefineVerbatimEnvironment{VerbatimWrap}{Verbatim}{
breaklines=true, 
breakindent=0pt, 
breaksymbol={},
fontsize=\scriptsize,
}

\title{Learning Scientific Exploration from Human Research Decision Trajectories}

\author{Xuchen Gong}
\author{Shane Gu}
\author{Haokun Liu}
\author{Dixi Yao}
\author{Chenhao Tan}
\author{Tian Li}
\affiliation{University of Chicago}

\metadata[Email]{\href{xuchengo@uchicago.edu}{xuchengo@uchicago.edu}}
\metadata[Code]{\href{https://github.com/xuchengong/researchtrails}{github.com/xuchengong/researchtrails}}
\metadata[Data \& Models]{\href{https://huggingface.co/researchtrails}{huggingface.co/researchtrails}}
\metadata[Data Visualizer]{\href{https://xuchen.io/researchtrails-page/visualizer/}{researchtrails.org/visualizer}
}
\metadata[Blog]{\href{https://xuchen.io/researchtrails-page/blog/}{researchtrails.org}}

\begin{document}

\abstract{
A key challenge in building AI systems for scientific research is enabling \textit{scientific exploration}: the systematic process of investigating unknown phenomena or ideas to gain new knowledge through sequences of research decisions and actions.
Yet this process is largely missing from existing scientific corpora; for example, research papers primarily record final outcomes rather than the trajectories that produced them. 
In this work, we introduce \textcolor{black}{\textbf{ResearchTrails}}, a dataset of \textcolor{black}{\textbf{human research trajectories constructed from Git repositories}}, where \textcolor{black}{\textbf{commit histories}} serve as proxies for research exploration. We develop an automated and scalable pipeline that extracts structured research trajectories from repository commits, capturing successive changes to methods, experiments, and ablations. We characterize the resulting dataset and show that these trajectories contain meaningful signals about intermediate research decisions beyond what final papers reveal. We further demonstrate utilities of ResearchTrails in multiple use cases, including retrieving human research experience as external skills at test time and training models on research trajectories to improve generalization to new research decisions. Our results suggest a path toward AI systems that learn not only from the products of science, but from the evolving process of discovery itself.
}
\maketitle



\section{Introduction}
\label{sec:intro}


Existing systems for automating scientific research powered with large language models (LLMs) have demonstrated promising capabilities across many components of the research process, including generating research ideas, synthesizing literature \citep{openscholar,skarlinski2024language}, writing code \citep{yang2024swe}, and executing other relevant well-specified tasks \citep{expbench,mlebench}. Agents have been chained together into agentic workflows to simulate the entire research lifecycle or simulate certain research communities \citep{yu2024researchtown}.


Despite rapid progress in automated research, how models can learn the evolving process of \textit{scientific exploration}, i.e., how to make a sequence of research decisions as a project unfolds, is underexplored. In addition to generating ideas or executing a predefined plan, humans typically perform a series of explorations during the research process, such as modifying methods after observing unexpected behavior, introducing new experiments to test hypotheses, and abandoning or revising directions that appear unpromising \citep{klahr1988dual,dunbar1994scientists}. 


A central bottleneck of improving LLM's capabilities in scientific exploration is the lack of supervision. Research papers remain the most popular way of sharing findings, but they primarily describe the final outcome of projects. The winding processes that produce the outcomes are absent \citep{drawer}. Recent works have explored agentic research trajectories by letting LLMs interact with synthetic research environments (e.g., other LLMs) \citep{cai2026ai,si2026towards}, or synthesizing reasoning traces \citep{wang2026learning}. However, these trajectories may not reflect the decision processes of human researchers, which is challenging to obtain at scale.
Therefore, we ask: 
\textit{Can we create human research-decision trajectories at scale? Does learning from those decision trajectories improve scientific exploration beyond learning only from final research outcomes?}


In this work, we introduce ResearchTrails, a dataset of human research trajectories constructed from the development histories of scientific projects. Our key insight is to use Git repositories as proxies of how human research evolves over time.  We develop an automated pipeline to transform commit histories into a series of research decisions in the exploration process. Daily commit histories are messy signals containing low-level operations such as updating an implementation, changing the training configuration, or switching to another dataset. ResearchTrials contain higher-level abstractions about research decisions on methods, experiments, and ablations.
This gives us a view of how a research project evolves before arriving at the final discoveries. Furthermore, we abstract project-agnostic research patterns from trajectory samples and observe insightful research strategies. We  study how ResearchTrials can be useful for improving AI-assisted scientific exploration through multiple use cases, including test-time and training-time methods. First, we use the trajectories as external knowledge in the form of demonstration data or  skills. Second, we explore supervised and reinforcement fine-tuning on research trajectories to explore the next research decision. We showcase that ResearchTrails enables models to learn from the sequence of scientific exploration steps. 


Our contributions are summarized as follows. In Section~\ref{sec:data}, we introduce ResearchTrails and a scalable pipeline for extracting evidence-grounded research-decision trajectories from Git repositories. In Section~\ref{sec:properties}, we show that these trajectories capture meaningful patterns of scientific exploration, including how methods, experiments, and ablations evolve over time. In Section~\ref{sec:applications}, we demonstrate that ResearchTrails can support scientific exploration as test-time knowledge through demonstrations and reusable skills, and provide training-time supervision through ordered research-decision sequences.

\begin{figure}
    \centering
    \includegraphics[width=\linewidth]{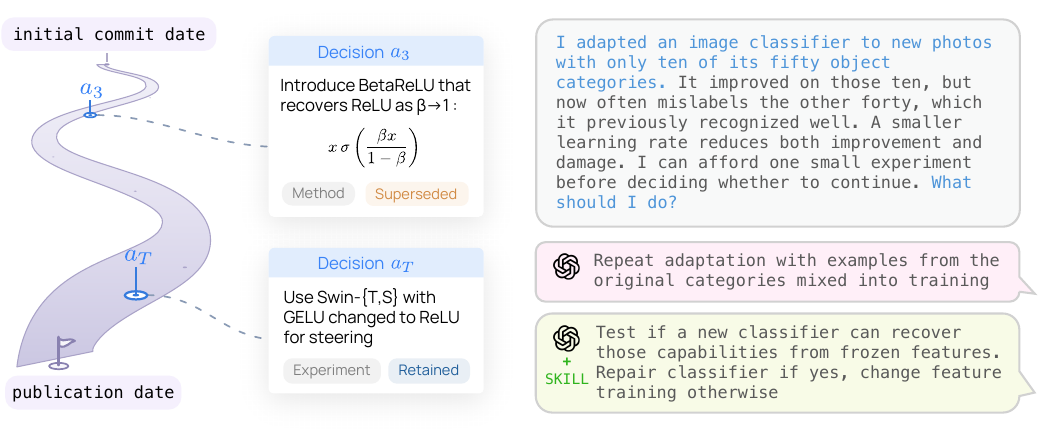}
    \caption{Overview of ResearchTrails: learning scientific exploration from human research trajectories. Left: We extract research decisions from GitHub commit history. Right: We utilize these trajectories to improve LLMs' capability of assisting with research exploration. For example, skills distilled from trajectories let GPT-5.6 Sol suggest more diagnostic and actionable interventions.}
    \label{fig:placeholder}
\end{figure}
\section{Related Work}


\textbf{Benchmarks for autonomous research.} Recent work has developed benchmarks for evaluating LLMs on different stages of scientific research, such as idea generation~\citep{si2025can,qiu2025ai}, hypothesis generation and evaluation~\citep{liu2025researchbench,liu2025hypobench,o2025sparks}, experimentation~\citep{huang2023mlagentbench,chan2025mle}, paper writing~\citep{wang2024autosurvey}, or paper reviewing~\citep{liu2023reviewergpt,idahl2025openreviewer}. Other benchmarks explore multiple tasks in broader research workflows, including executing concrete scientific tasks~\citep{chen2025scienceagentbench,du2026deepresearch,speedrunning,paperbench} and evaluating  AI agents on open-ended machine learning  research~\citep{chen2026mlr,nathani2025mlgym}. These benchmarks increasingly capture realistic components of scientific research, but they primarily evaluate task completion or quality based on the final output rather than how research decisions evolve over the course of a project.

\textbf{Methods and systems for research agents.}
Beyond benchmarking, recent systems have developed new tools, algorithms, and pipelines to automate the general research process. For example, \citet{shao2025dr} build agentic systems that integrate web browsing, document retrieval, and reasoning to produce answers and reports. 
Other work improves different part of the research process such as idea generation~\citep{baek2025researchagent,li2025chain} or using scientific skills~\citep{kassis2026scientific,claudescientificskills2026,zhang2026vibe}. Extensive works have built end-to-end research automation pipelines or simulated research communities~\citep{lu2024ai,yamada2025ai,schmidgall2025agent,tang2025ai,gottweis2025towards,yu2024researchtown}. 
Despite advancements of these approaches, their research trajectories are largely generated by LLMs or induced by predefined agentic workflows, rather than learned from the development histories of human researchers.

\textbf{Human research artifacts.} Research papers, though often only record the final research outcome, have been widely used to provide training signals for research agents \citep{goel2025training,he2026giants,liu2026sci}. Several recent AI scientist systems explicitly incorporate humans into the research loop, allowing humans to provide feedback for LLMs~\citep{gottweis2026accelerating,swanson2025virtual}. Concurrent work \citep{chen2026repo} distills files in the  GitHub repositories into LLM skills, whereas we focus on specific repositories with rich development histories and study their commit histories as evidence of how human research decisions evolve.

\textbf{Learning from process traces.} A growing line of work uses intermediate process traces to provide supervision for various tasks. ScholaWrite \citep{le2025scholawrite}  uses LaTeX edits to improve AI writing, and SWE-RL \citep{wei2026swe} mines pull requests from open-source software to help reproduce developers' patches. In contrast, we focus on scientific exploration, and we aggregate many low-level commits into research decisions about methods, experiments, and ablations. 

\section{ResearchTrails: From Git Histories to Research Trajectories}
\label{sec:data}

In this section, we present the formulation of research decision trajectories, introduce how such data can be extracted from existing GitHub repositories at scale, and present our workflow in detail. 

We consider the \textit{research exploration} problem where an LLM system assists with insightful discussion on how to proceed with the existing research project. Given a research query on what has been attempted on a research question, the AI-assisted workflow suggests the next research decision that is promising. Specifically, it can propose a change in the project development, such as `add load balancing loss to the routing objective', `evaluate model performance after being $10\times$ deeper', or `visualize the loss landscape'. 

The desirable data for this task is the process of human research explorations, which unfortunately is typically not presented in a paper. However, our insight is that the GitHub repository associated with a paper can maintain historical commits that reflect a long period of explorations. Individual commits can be too low-level to represent meaningful research decisions, whereas a sequence of commits and code differences can collectively correspond to higher-level research decisions. For example, the commit sequence
$$\small
\begin{aligned}
\textsf{change router implementation},\;\;
\textsf{update the training bash script},\;\;
\textsf{visualize experiment results}
\end{aligned}
$$
yields the decision `study new router as an alternative to the original routing method'.
\looseness=-1

Therefore, we extract a research decision sequence $(a_1,a_2,\ldots,a_T)$ (a research trajectory) from the commit sequence. In this work, we focus on repository-observable research decisions that can be systematically grounded in repository evidence. We formulate each $a_t$ as a scientific change that meaningfully alters a project's proposed method, experimental design, or ablation study. 

\begin{wrapfigure}{R}{0.42\textwidth}
\centering
\includegraphics[width=\linewidth]{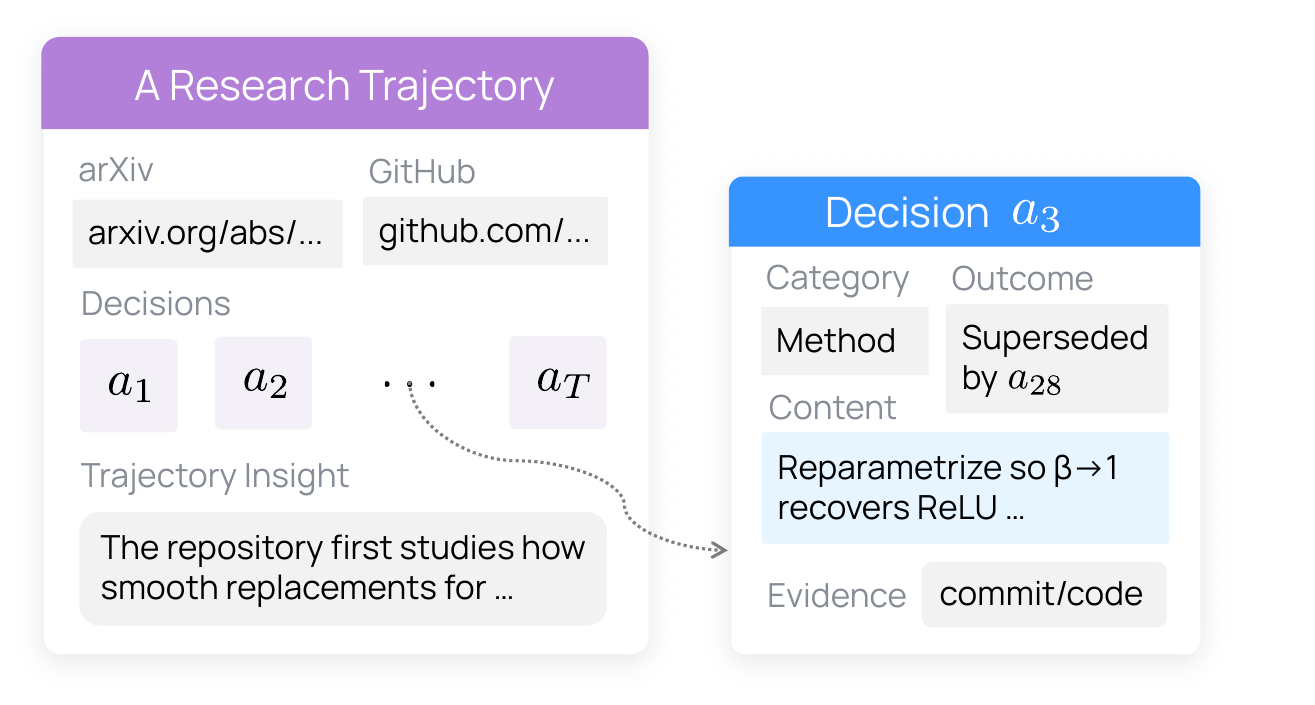}
\caption{Structure of a research trajectory (left) and a research decision (right).}
\label{fig:decision}
\vspace{-0.5cm}
\end{wrapfigure}
Multiple commits may be grouped into one decision when they correspond to the same underlying scientific change, and one commit can reflect multiple decisions. In addition, we separate research decisions from ordinary engineering changes: formatting changes, dependency updates, typo corrections, and file differences from model weights or notebook metadata are generally not treated as research decisions.

Each research decision is represented in a data structure containing textual descriptions of decision \texttt{content}, together with commit ID references and code evidence that supports the annotation (Figure \ref{fig:decision}). In addition, each decision has two property labels describing its \texttt{category} and \texttt{outcome}. A \texttt{category} label is either \textit{method}, which builds or revises the project's contribution itself; \textit{experiment}, which changes how the contribution is evaluated; or \textit{ablation}, which studies the sensitivity of the components of the proposed method. An \texttt{outcome} label is either \textit{\textcolor{retained}{retained}}, which remains in effect at the end of the observed history; \textit{\textcolor{superseded}{superseded}}, which a later decision replaces in the same research role; or \textit{\textcolor{abandoned}{abandoned}}, which is removed without a clear successor. We present the instructions for labeling decisions in Appendix \ref{sec:app:property}.

\begin{figure*}[t]
\centering
\includegraphics[width=\textwidth]{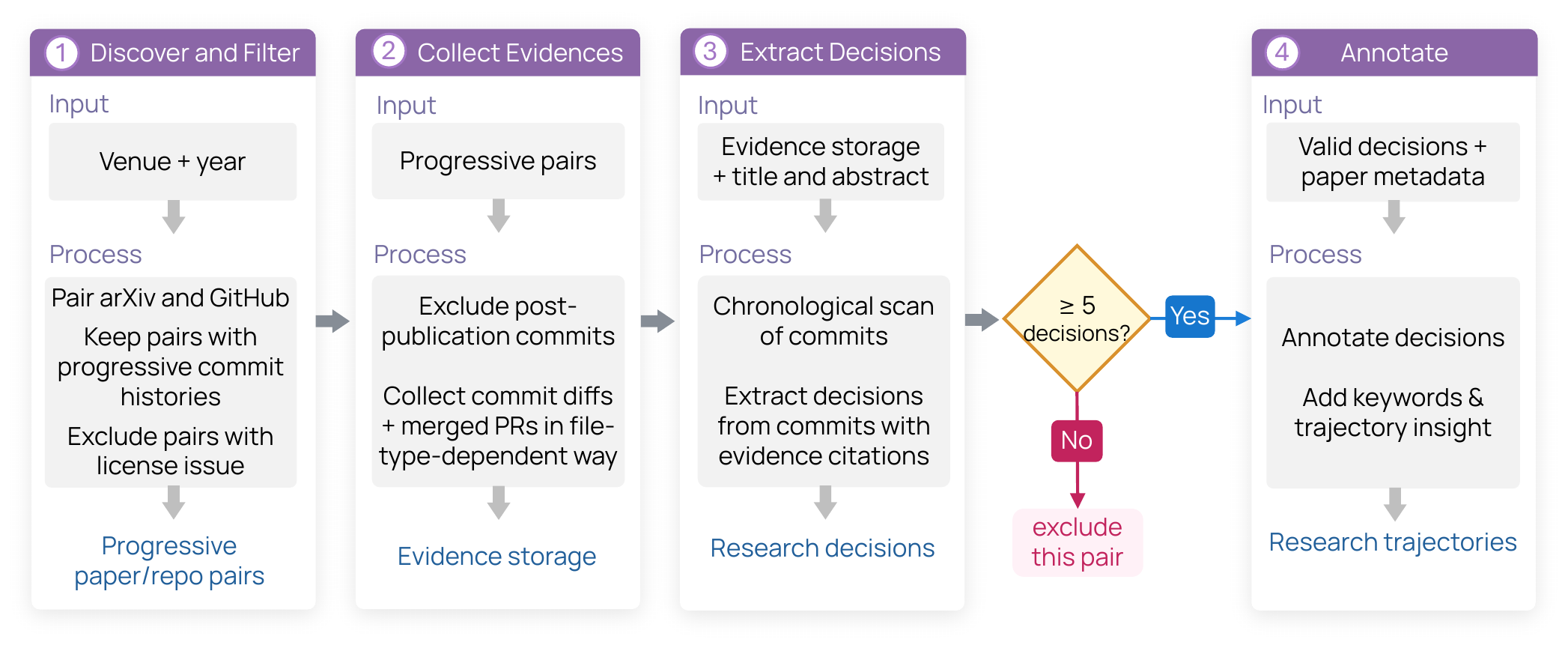}
\vspace{-1em}
\caption{Workflow of extracting research decisions from GitHub repositories. Candidate paper/repository pairs are discovered and filtered before collecting repository code evidence. An LLM then processes that evidence chronologically, merges commits into decisions, with citations checked against the stored Git evidence. Only repositories with at least five validated decisions are annotated.}
\label{fig:workflow}
\end{figure*}

We present our workflow for automatically extracting decisions from commit histories in Figure \ref{fig:workflow}, which is composed of four steps: (1) discover the paper/repository pairs for a given vanue-year and filter out pairs whose repository has no progressive commits; (2) collect code differences in commits as evidence excerpts; (3) extract decisions from code evidences, and (4) annotate the decisions following the requested data structure. The details of each step are described in Section \ref{sec:data:filter}--\ref{sec:data:annotate}, with further implementation designs in Appendix \ref{sec:app:property}.

\subsection{Selecting Repositories with Observable Research Development} \label{sec:data:filter}
In the first stage, we identify pairs of  papers and repositories with observable progressive prepublication development history. 
Specifically, given a venue and year, we retrieve the paper list from Semantic Scholar \citep{semanticscholar} and retain paper/repository pairs with progressive commit histories before entering the next steps. First, we resolve and pair the arXiv and GitHub links: For each paper with an arXiv identifier, we collect all the GitHub repository links from the arXiv PDF text, excluding References, Bibliography, or Appendix, so that links appearing only in the supplementary material are not harvested. Next, a candidate repository is retained only if its README contains the paper's full title. We then count the number of commits before the arXiv date and record the number of days between the first commit and the arXiv date. We exclude paper/repo pairs with fewer than 5 pre-publication commits or fewer than 30 span days between the first commit date and paper publication date.

We additionally apply the following filtering rules to exclude repositories whose code mostly contains content irrelevant to the paper's research decisions: We exclude known repositories that maintain several papers' commits in one repo; we exclude papers with fewer than 2 pre-publication commit messages that match a pool of research-related keywords; we exclude papers whose GitHub license restricts code usage for research purposes. Pairs surviving every filtering rule are labeled \emph{progressive}.

\subsection{Evidence-Grounded Trajectory Extraction} \label{sec:data:extract}

In the extraction stage, we turn a commit history into a research trajectory under two principles. First, one decision may take many commits to implement, and one commit may implement several decisions. Their mapping is \textit{many-to-many}, and we present our method to scan over commits below. Second, we enforce \textit{annotation reliability} by grounding every decision with repository evidence in the form of code pieces.

\paragraph{Evidence storage.} For each progressive pair, we create a partial repository clone that omits file contents larger than a specified size threshold, and walk its default branch up to the end of the paper's arXiv upload month. Commits are enumerated following the commit time. Where the GitHub API associates merged pull requests (PR) with default-branch commits, the PR commits are interleaved at their merge anchors so that work developed on a branch appears at the point it entered the history.

We record commit contents into an `evidence storage' via the following rules: (1) Informative file types such as source, configuration, and notebook files are \emph{read at creation}. For the first commit, we store the file's complete text up to a character cap; for every later commit changing that file, we record only a bounded diff. (2) Notebooks are parsed to retain only the code input, removing large spurious diffs produced by simply re-running a notebook. (3) For artifacts such as model checkpoints and logs, we only store metadata such as path and change status.


\paragraph{Mapping commits to decisions.}
A \textit{decision} expresses exactly one scientific choice regarding the \textit{method}, \textit{experiment}, or \textit{ablation}. Setup code, loaders, logging, and reporting utilities are merged into the choice they implement. In this step, we call LLM APIs to extract decisions from commit histories. LLMs read the stored evidence file together with the title and abstract of each paper only, which are references to assist with determining, for example, whether a commit is research-relevant or a side-product of a library upgrade.

Each LLM call receives the earlier decisions before the current batch of commits, based on which it classifies the role of each commit: 
\textit{irrelevant}, \textit{support} (adds evidence to an existing decision), \textit{revise} (changes one decision), \textit{new} (begins one), \textit{multiple} (contains several), or \textit{uncertain}. An uncertain commit is stored in an evidence-backed \textit{open thread}. An open thread may later be merged into an existing decision, promoted to being a decision together with other commits, or rejected if never merged or promoted. An example of mapping a commit to decisions is presented in Table \ref{tab:commit_assessment_exapmle}.

\paragraph{Grounding decision annotations with repository evidence.}
If a commit is classified as supporting one or multiple decisions, its ID will be added to those decisions' evidence list. Then the LLM response contains assessments of every commit in the batch, created or revised decisions, created or revised open threads, thread resolutions, and decision merges. If the response omits any commit's assessment or names a decision unsupported by evidence, we discard the entire response and retry the commit under a halved input budget. After obtaining the commit assessments, the harness applies this delta to the complete decision trajectory, unioning commit identifiers and evidences. 

\subsection{Verification and Trajectory Annotation} \label{sec:data:annotate}
After all commits have been accessed, the LLM validates every claimed commit identifier, file path, date, and code difference excerpt against the evidence storage. An excerpt that is not an exact substring of a stored patch is discarded, and a decision with no valid evidence citation is rejected. Each decision records its lifecycle outcome (\textit{retained}, \textit{abandoned}, or \textit{superseded}), where a superseded decision must name its successor; a successor link whose target does not survive validation is dropped.

Accepted decisions are ordered by their first cited commit date. A repository with fewer than five validated decisions is dropped. Otherwise, the LLM performs one final call to derive keywords and \textit{trajectory insight}, a trajectory-level observation from the validated decisions alone.

\section{Dataset Properties and Research Trajectory Patterns}
\label{sec:properties}

In the following, we discuss the dataset properties in Section \ref{sec:method:stats} and present several general research patterns extracted from the annotated research trajectories in Section \ref{sec:method:insight}.

\subsection{Dataset Overview}
\label{sec:method:stats}

Utilizing the workflow in Section \ref{sec:data}, we process 15K papers from NeurIPS 2022--2025 due to this venue's relatively broad topic coverage and abundant indices of arXiv links on Semantic Scholar. We find 5K out of the 15K indexed papers have a matched repository. Next, the filtering stage retains 677 progressive pairs (13\%), and the final dataset contains \textbf{599} paper/repository pairs with annotated human research trajectories, which consist of \textbf{13,271}   research decisions. This showcases that out of all the collected papers, $\sim$4\% of them can yield a progressive research decision trajectory. 
\begin{figure*}[t]
\begin{minipage}{0.75\textwidth}
\centering
\small
\setlength{\tabcolsep}{2.2pt}
\begin{tabular}{llcccc}
\toprule
\multirow{2}{*}{Stage} & \multirow{2}{*}{Outcome} & \multicolumn{4}{c}{Count} \\
\cmidrule(l){3-6}
 &   & 2022 & 2023 & 2024 & 2025 \\
\midrule
Input   & Papers indexed for the year    & 2790 & 3587 & 4544 & 4313 \\
\midrule
Stage 1 & No arXiv identifier                  & 558 & 693 & 862 & 934 \\
      & No repository link found             & 1142 & 1477 & 1779 & 1670 \\
      & README does not name paper       & 214 & 237 & 330 & 321 \\
      & \cellcolor{lightpurple!40}\textbf{Retained pairs}              & \cellcolor{lightpurple!40}\textbf{845} & \cellcolor{lightpurple!40}\textbf{1198} & \cellcolor{lightpurple!40}\textbf{1594} & \cellcolor{lightpurple!40}\textbf{1400} \\
\midrule
Stage 2 & Fewer than 5 total commits            & 190 & 235 & 333 & 340 \\
      & Fewer than 5 pre-paper commits       & 279 & 420 & 499 & 394 \\
      & Pre-paper span below 30 days         & 187 & 282 & 413 & 380 \\
      & Fewer than 2 research commits        & 72 & 92 & 116 & 112 \\
      & \cellcolor{lightpurple!40}\textbf{Progressive pairs}           & \cellcolor{lightpurple!40}\textbf{111} & \cellcolor{lightpurple!40}\textbf{168} & \cellcolor{lightpurple!40}\textbf{229} & \cellcolor{lightpurple!40}\textbf{169} \\
\midrule
Stage 3 & Stored commits                       & 16,661 & 25,345 & 43,664 & 41,646 \\
      & Stored evidences                     & 104,981 & 173,469 & 2,759,132 & 514,430 \\
\midrule
Stage 4 & Fewer than five decisions            & 16 & 12 & 22 & 5 \\
      & \cellcolor{lightpurple!40}\textbf{Annotated pairs}             & \cellcolor{lightpurple!40}\textbf{90} & \cellcolor{lightpurple!40}\textbf{150} & \cellcolor{lightpurple!40}\textbf{204} & \cellcolor{lightpurple!40}\textbf{155} \\
      & \cellcolor{lightpurple!40}\textbf{Annotated decisions}         & \cellcolor{lightpurple!40}\textbf{2176} & \cellcolor{lightpurple!40}\textbf{3593} & \cellcolor{lightpurple!40}\textbf{5260} & \cellcolor{lightpurple!40}\textbf{2242} \\
\bottomrule
\end{tabular}
\label{tab:neurips-funnel}
\end{minipage}
\hfill
\begin{minipage}{0.24\textwidth}
\vspace{0.4cm}
\centering
\includegraphics[width=\linewidth]{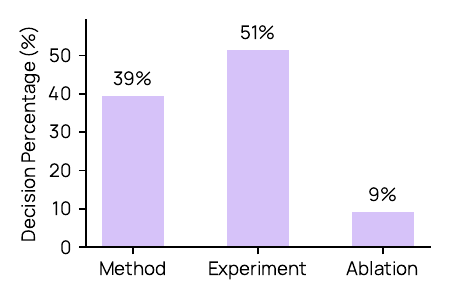}\\
\includegraphics[width=\linewidth]{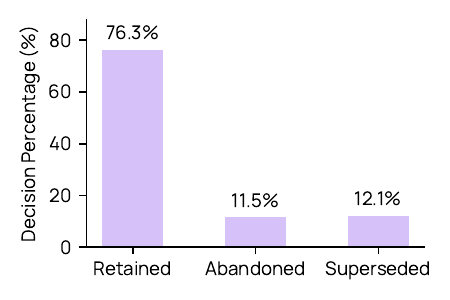}\\
\includegraphics[width=\linewidth]{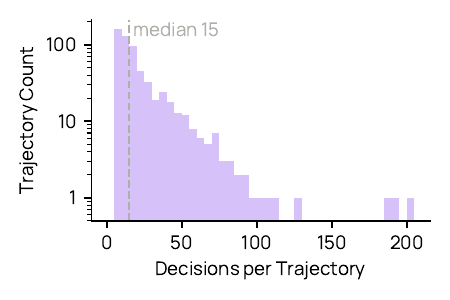}
\label{fig:statistics}
\end{minipage}
\vspace{-0.5cm}
\captionof{table}{Count of paper/repository pairs throughout the workflow for NeurIPS 2022--2025 (left) and summaries of other dataset statistics (right), including the distributions of decision \texttt{category}, decision \texttt{outcome}, and trajectory lengths.}
\end{figure*}

Table~\ref{tab:assessments} in the appendix reports the distribution of commit accesses, where 57\% of commits are \textit{irrelevant}: In repositories showing meaningful commit activities before the publication date, roughly one commit in three carries a research decision. 

For our annotated 13,271 decisions, they comprise 5226 changes in methods (39\%), 6826 in experiments (51\%), and 1219 in ablations (9\%). Among these, 76.3\% are retained, 11.5\% are abandoned, and 12.1\% are superseded. Decision counts per trajectory range from 5 to 205, with a median of 15, and 35\% of all the trajectories have $\geq 20$ decisions. A decision has 7.5 evidence citations on average.

\subsection{How Research Trajectories Evolve} \label{sec:method:insight}
\begin{figure}[t]
\centering
\includegraphics[width=\textwidth]{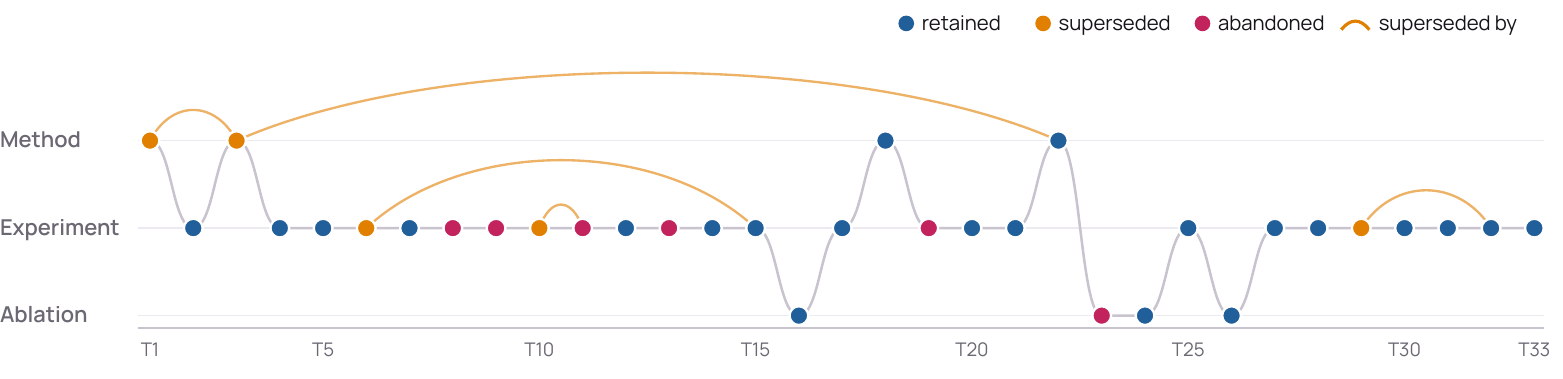}
\caption{Annotation example for \citet{hu2026curvature}. It plots a trajectory by category, showing how a project moves between building its method, testing it, and ablating it. Orange arcs connect a superseded decision to the one that replaced it.}
\label{fig:example3}
\end{figure}

In this section, we examine temporal structures of annotated research trajectories and reveal recurring research-decision patterns across projects. Figure \ref{fig:example3} presents an annotation example, which display winding research processes with back-and-forth development of methods, experiments, and ablations. We present more examples with detailed decision contents in \ref{fig:example1}--\ref{fig:example6} in Appendix \ref{sec:app:property}. \looseness=-1 

\begin{figure}[t]
\centering
\includegraphics[width=\linewidth]{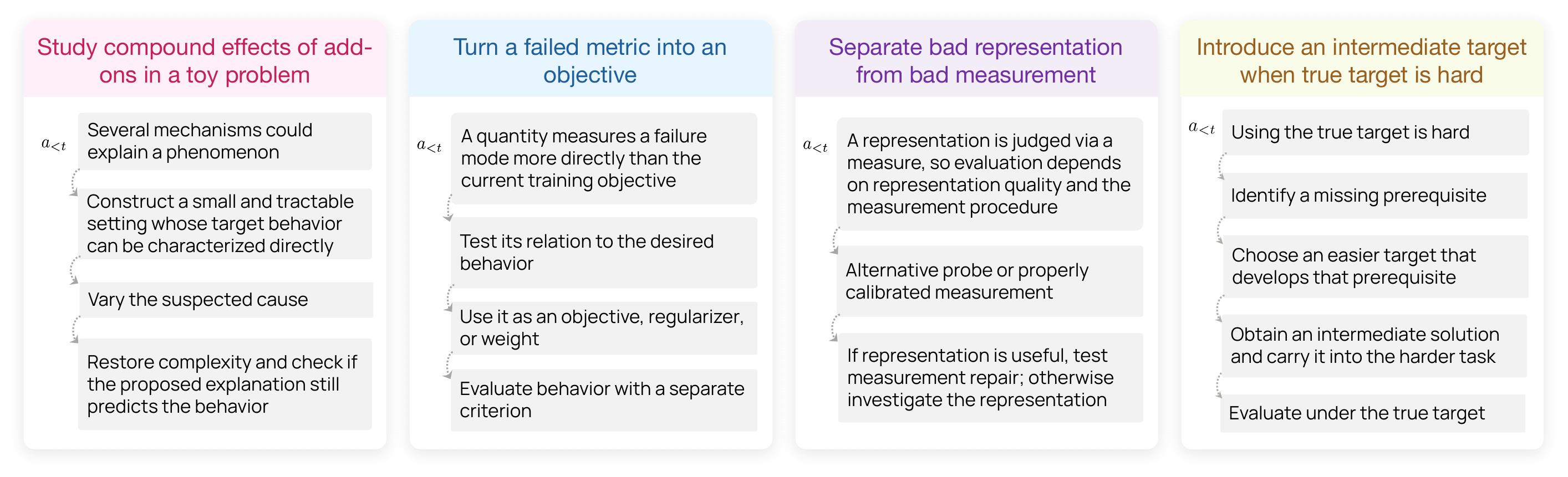}
\vspace{-0.7cm}
\caption{Recurring research patterns extracted by GPT-6 Astra that generalize across research projects.}
\label{fig:codex_insight}
\end{figure}

While research trajectories are project-specific, we observe common research decision structures across projects. 
We abstract these structures into research patterns, each comprising a project state, informed by the preceding decisions ($a_{<t}$), and a sequence of subsequent research actions. Figure~\ref{fig:codex_insight} presents four such patterns and illustrates two shared characteristics.


First, the patterns organize research decisions around abstract project states, as opposed to focusing on specific project details. For instance, one shared pattern is to incorporate the desired properties into the objectives (the second pattern in Figure~\ref{fig:codex_insight}).   
Since patterns are generalizable across projects, they can be given to a language model as guidance or skills for choosing a next decision, which we examine in Section~\ref{sec:applications}. 

Second, 
several patterns prioritize diagnostic investigation over blindly changing to different methods. For example, when multiple compounded mechanisms  explain an observed phenomenon, the first pattern suggests to construct a small and tractable problem to isolate the causes. The third pattern suggests to use an alternative probe or a calibrated measurement procedure to distinguish limitations of the representation from limitations of its evaluation. These examples illustrate how research patterns can suggest targeted investigations that inform subsequent research choices. This motivates encoding the patterns as  skills for language models (Figure~\ref{fig:user_simulation} in Section~\ref{sec:applications}).

\section{Applications}
\label{sec:applications}
Given an instruction prompt $p$ and the previous human research decisions $a_{<t}=(a_1,\ldots,a_{t-1}),$ we denote the input to LLM as $h_t=(p,a_{<t}).$ Our goal is to obtain an improved agentic workflow that can assist with human research exploration, i.e., yielding a promising next decision $\hat a_t\sim\pi_\theta(\cdot| h_t)$ that leads to observations that expand the boundaries of human knowledge. 


\textbf{Evaluation Metrics.} Since it remains an open question to evaluate the effectiveness of an arbitrary decision $\hat a_t$ \citep{si2026ideation,researchgym}, in this work, we propose using held-out real human research trajectories that we collected as user simulations, and we measure if the generated next decision is strategically similar to the actual decision. We also present case studies showing the improved quality of the generated next decision. 
We use an LLM judge \citep{zheng2023judging,liu2023g} to evaluate the similarity between two research decisions via scoring three aspects of a research decision independently with \{0,1,2\}: \textit{component match} (what part of the project is acted on, such as baselines, evaluation datasets, model architecture, etc.), \textit{operation match} (the scientific action, such as introduce, extend, replace, remove, extend, etc.), and \textit{specification match} (concrete choices). We then define \texttt{score} as the sum of the match in \textit{component} and \textit{operation}, discarding \textit{specification} since, for example, `increase batch-size to 16' and `increase batch-size to 32' are strategically the same research decision. We present the judge rubric in Appendix \ref{sec:app:application} and examples in Table \ref{tab:rubric_example}. We use judge GPT-5.6 Sol and Opus 5, whose judge score variance and judge quality are evaluated in Appendix \ref{sec:app:application}. \looseness=-1


In the following, we propose both test-time and training-time approaches to improve decision similarity scores on held-out human research trajectories. For test-time approaches (Section~\ref{sec:exp:test}), we derive harness by (1) using a few trajectories as demonstration data or (2) distilling skills from trajectories, which are compared with (3) distilling skills from final papers only. For the training-time approach (Section~\ref{sec:learning}), we train an autoregressive policy over the next decision $\pi_\theta(a_t| h_t)$ via supervised and reinforced fine-tuning.

\subsection{Harness: Use Trajectory Examples as Demonstrations and Skills}
\label{sec:exp:test}

\begin{figure}[t]
\begin{minipage}{0.28\textwidth}
\centering
\includegraphics[width=\textwidth]{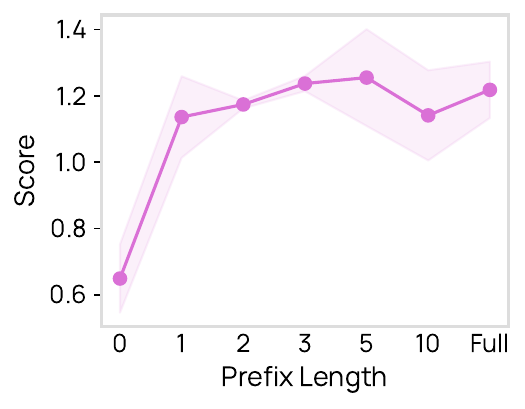}
\caption{Performance plateaus with approximately 3–5 prior decisions in context.}
\label{fig:prefix_length}
\end{minipage}
\hfill
\begin{minipage}{0.66\textwidth}
\centering
\small
\begin{tabular}{@{}llll@{}}
\toprule
Method  & Prefix-1 & Prefix-2  &  Prefix-3  \\
\midrule
Baseline  & 1.256 $\pm$ .090 & 1.289 $\pm$ .022 & 1.445 $\pm$ .063 \\
\textcolor{myyellow}{All demos}  &  1.328 $\pm$ .095 & 1.316 $\pm$ .068 & 1.373 $\pm$ .071  \\
\textcolor{mypink}{Random two demos} & 1.241 $\pm$ .037 & 1.216 $\pm$ .020 & 1.274 $\pm$ .121 \\ 
\textcolor{mygreen}{Retrieve two demos} & 1.361 $\pm$ .043 & 1.460 $\pm$ .016 & 1.498 $\pm$ .080  \\ 
\textcolor{myyellow}{Shuffled skill} & 1.363 $\pm$ .045 & 1.308 $\pm$ .182 & 1.391 $\pm$ .053 \\
\textcolor{myyellow}{Final paper skill} & 1.500 $\pm$ .020 & 1.231 $\pm$ .064 & 1.271 $\pm$ .034 \\
\textcolor{mygreen}{Skills} & \textbf{1.697} $\pm$ .088 & \textbf{1.557} $\pm$ .089  & \textbf{1.607} $\pm$ .009 \\
\bottomrule
\end{tabular}
\caption{Decision similarity of the predicted next decision based on previous $\{1,2,3\}$ decisions. Sores are judged by GPT-5.6 Sol and are averaged across evaluation trajectories, each with 3 runs; Scores judged by Opus 5 are in Table \ref{tab:harness_opus}.}
\label{tab:harness}
\end{minipage}
\end{figure}

To understand the difficulty of predicting the next decision, we vary the number of prior decisions provided in context and compare the scores obtained by predictor Gemini 3.1 Flash-Lite. We observe in Figure \ref{fig:prefix_length} that performance saturates when the number of prior decisions $a_{<t}$ is around 3–5, indicating that the next decision becomes less uncertain with abundant decision history, such as at a late stage of a project. 



To focus on the most challenging scenario where decision history is short and the next decisions are uncertain, we predict the next decision using prefixes in early timesteps of lengths $\{1,2,3\}$. We utilize 19 training repositories' trajectories for training and 134 held-out repositories for evaluation. We use Gemini 3.1 Flash Lite to predict the next decision based on the previous trajectory and compare the following methods: (1) \textbf{Baseline:} directly prompting the model; (2) \textbf{All demos:} use all the training trajectories as demonstrations \citep{demonstration}; (3) \textbf{Random two demos:} randomly choose two training trajectories as demos; and (4) \textbf{Retrieve two demos:} select two most relevant training trajectories as demos; (5) \textbf{Skill:} use skills distilled from the training trajectories by Opus 5; (6) \textbf{Shuffled skill:} use skills distilled from the same trajectories with the temporal order randomly permuted within each project, which helps evaluate if the order of decisions is important; (7) \textbf{Final paper skill:} use skills distilled from the final paper content only, which helps evaluate if intermediate research decisions are important.\looseness=-1

We observe that experience from other projects effectively helps predict the next decision when a project has only 1-3 decisions so far. The two most effective methods are `Retrieve two demos' and `Skills': (i) Retrieval outperforms random selection by 0.196 and using all demos by 0.101 on average, showing that the most relevant demonstrations help with predictions, and additionally using irrelevant demonstrations distracts and harms performance. Furthermore, (ii) Skills perform the best, outperforming retrieval by 0.181, final paper skills by 0.286, and baseline by 0.290 on average. This indicates that learning from those decision trajectories improves scientific exploration beyond learning only from final research outcomes. 

\begin{figure}[t]
    \centering
    \includegraphics[width=\linewidth]{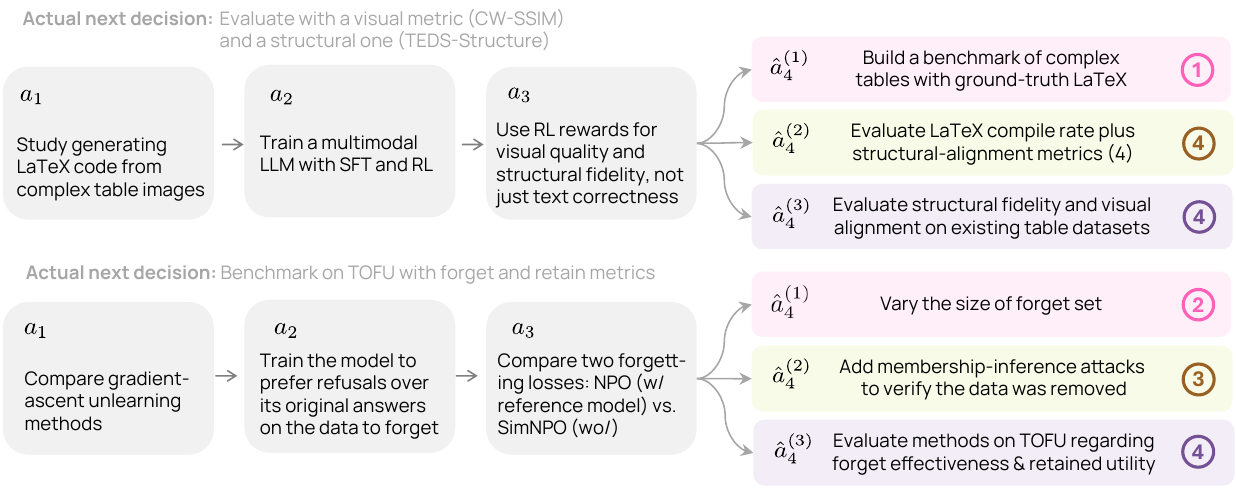}
    \caption{Sampling three next decisions based on previous decisions from held-out real trajectories. Upper: The decisions are generated with method `Skills'; Bottom: The decisions are generated with `Retrieve two demos'. The circled values after generated decisions are the decision similarity scores.}
    \label{fig:generation}
\end{figure}

\begin{figure}[t]
    \centering
    \includegraphics[width=0.9\linewidth]{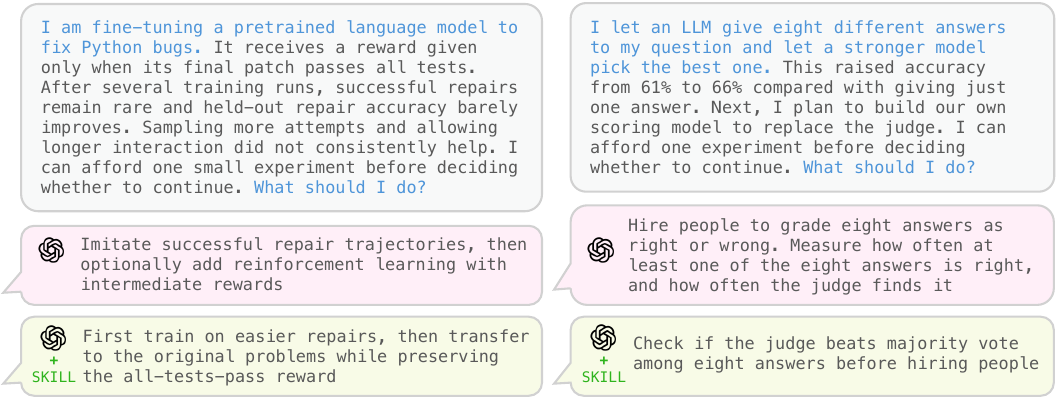}
    \caption{User simulations where a user asks GPT-5.6 Sol (high reasoning) research queries. In both examples, the skills let Sol suggest smaller and more interpretable interventions, changing the suggestion from \textbf{trying a remedy} to \textbf{determining which component needs repair}.}
    \label{fig:user_simulation}
\end{figure}

For a qualitative study, we present the generated next decisions and their scores on held-out real trajectories in Figure \ref{fig:generation}. We observe that the predictor can yield various decisions via sampling, and the generated decisions are generally of good quality with proper `harnesses' such as `Skills' and `Retrieve two demos'. Furthermore, we run user simulations where a researcher asks GPT-5.6 Sol research queries (Figure \ref{fig:user_simulation}), observing that research patterns used as skills let Sol suggest smaller and more interpretable interventions. 

\subsection{Training: Autoregressive Decision Modeling and RL}
\label{sec:learning}
For the training-time approach, we present a two-stage pipeline: autoregressive decision modeling (SFT cold start), followed by GRPO \citep{shao2024deepseekmath} with decision similarity rewards. In the SFT stage, we fine-tune $\pi_\theta(a_t| h_t)$, a Qwen3-8B model \citep{yang2025qwen3}, to predict the next decision via 
$$ \mathcal{L}_{\mathrm{SFT}}(\theta) = -\mathbb{E}_{(p,a_{1:T})\sim\mathcal{D}_{\mathrm{train}}}
\left[ \sum_{t=1}^{T} \log \pi_\theta \left( a_t | p,a_{<t} \right) \right]. $$
where $a_t$ is a sequence of tokens describing the decision content. Suppose it has length $|a_t|=S$, then $\log \pi_\theta ( a_t | p,a_{<t} ) = \sum_{s=1}^S \log \pi_\theta ( a_{t,s} | p,a_{<t})$. This objective learns the distribution $p(a_{1:T}| p)$, which represents plausible ways that a human research project may evolve given an initial instruction.

\begin{figure}[t]
    \centering
    \includegraphics[width=0.66\linewidth]{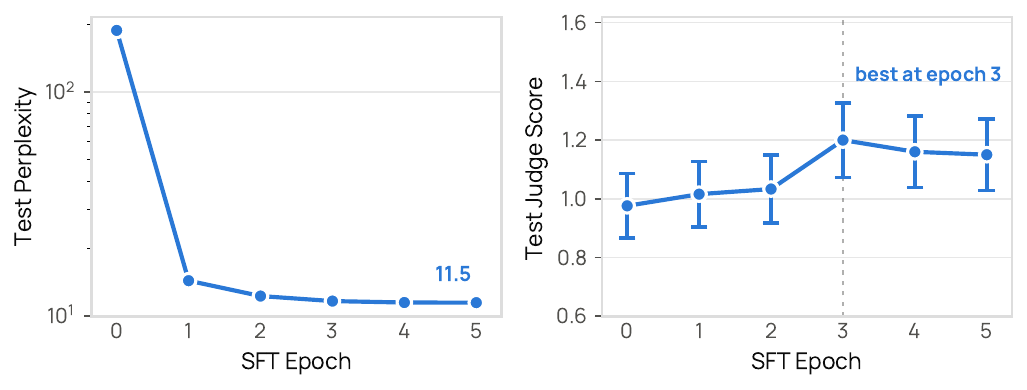}
    \includegraphics[width=0.32\linewidth]{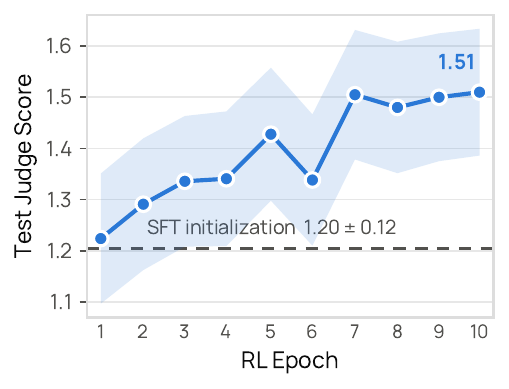}
    \caption{Test performance of two-stage training, including SFT cold-start (left and middle) and RL with next-decision rewards (right). We observe that SFT improves the probability of the next decision (left) whereas the generation quality plateaus at epoch 3, and the RL stage further improves generation quality (right). The error bars and shades indicate the score's 95\% confidence interval. Scores judged by Opus 5 are in Table \ref{tab:training_cross_judge}.}
\label{fig:training}
\end{figure}

We observe in Figure~\ref{fig:training} (left and middle) that likelihood and generation quality diverge during SFT. Test perplexity continuously falls, whereas the judge score of greedy predictions peaks at epoch 3 and does not improve further. Further gains therefore require optimizing generation quality directly.

Therefore, during RL, we use the SFT checkpoint as the frozen reference policy $\pi_{\mathrm{ref}}$. For each trajectory prefix, we sample $K$ decisions $\{\hat a_{k}\}_{k=1}^{K}$ from the current policy. The decision similarity judge assigns the component $c_{k}$ and operation match $o_{k}$ against $a$, yielding reward $R_{k}=c_{k}+o_{k}$ if $\hat a_{k}$ is a valid generation, and 0 otherwise. We optimize the policy with Dr.~GRPO \citep{liu2025understanding}, observing that RL effectively further improves the judge score from 1.20 to 1.51 (Figure~\ref{fig:training}, right).

\begin{figure}[t]
    \centering
    \includegraphics[width=\linewidth]{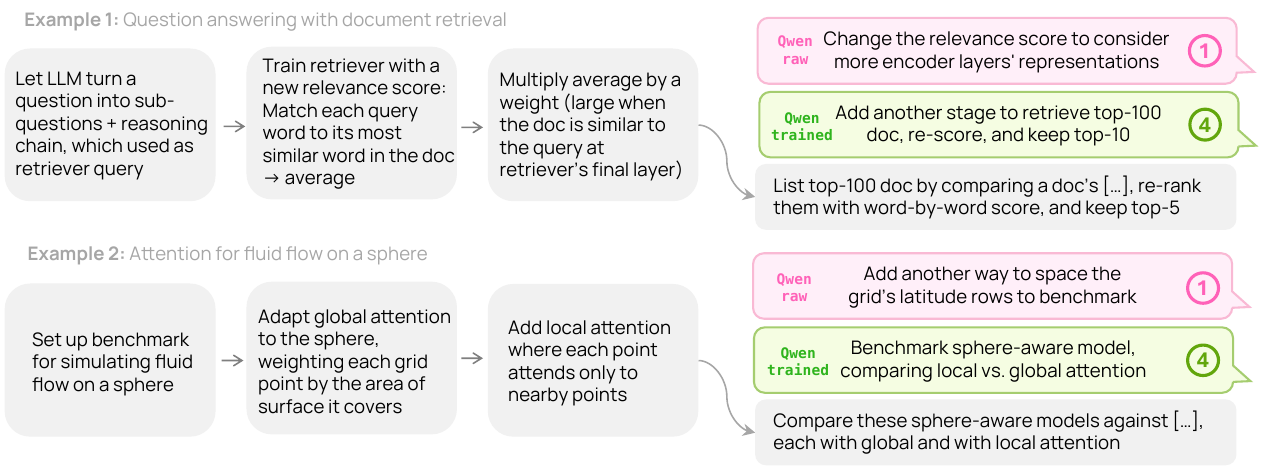}
    \caption{Generated next decisions from Qwen3-8B before and after our two-stage training. The examples are from held-out real trajectories, with grey blocks indicating the annotated decisions. The circled values are decision similarity scores for the corresponding generated decisions.}
    \label{fig:rl_user_simulation}
\end{figure}

To qualitatively compare the Qwen3-8B models before and after our two-stage training, we present their generated decisions on held-out real trajectories (Figure \ref{fig:rl_user_simulation}). We observe that training makes the model capable of predicting the next actions that are natural and logical next steps for the project.

\section{Conclusion} 


We introduced ResearchTrails, a dataset of human research decision trajectories extracted from the GitHub commit histories. These trajectories record what final papers omit, such as the abandoned ideas of proposed methods, omitted experiments, and ablations that shaped the final contribution. We observe and distill generalizable research skills from our trajectories. At test time, skills distilled from our trajectories improve frontier models' capability of generating the next decision on held-out projects, outperforming skills distilled from the final papers alone. At training time, we fine-tune Qwen3-8B via SFT and reinforcement learning with decision-similarity rewards, showing that our workflow is a scalable approach for providing learning supervision for scientific exploration.

We consider two directions for future work. First, we have focused on artificial intelligence conferences, while code histories can be maintained in other scientific research areas such as biology and physics. Second, our dataset only contains projects with Git commits, while there are other platforms containing research artifacts that can reflect research exploration patterns.

\clearpage
\bibliography{cite}
\bibliographystyle{iclr2027_conference}

\clearpage
\appendix
\section{Dataset Details} \label{sec:app:property}
In the following, we further detail the design considerations and implementations of our workflow for extracting research decisions from GitHub repositories. We use GPT-5.6 Sol (xhigh reasoning) for annotating decisions and writing trajectory insights.

\paragraph{Handling repository forks.} A repository fork inherits commits its authors did not write, and those commits should not be counted as the paper's research history. However, some type of fork is an exception: a personal repository forked into a laboratory organization after the work is complete has a creation timestamp that postdates every research commit, and windowing by that timestamp would discard the entire genuine history. We therefore distinguish the two cases: if the \emph{parent's} README also contains the paper's full title, the parent is the authors' earlier repository and the whole history is theirs; otherwise the parent is unrelated upstream, and every subsequent measurement counts only commits at or after the fork's creation timestamp.

\paragraph{Evidence storage.} For each file change in one commit, we record it as one evidence piece. If that file's diff is longer than 16,000 characters, it is split into multiple pieces. We turn commit contents into a storage of pieces (named `evidence storage') by the following rule:
\begin{itemize}[leftmargin=*]
  \item {Source, configuration, and notebook files} are \emph{read at
    creation}. The first commit that adds a path stores the file's complete
    text up to a character cap, which depends on the file type: 50K for Python, other source, and notebooks; 20K for shell and configuration. Every later commit changing that file records only a bounded diff.
  \item {Notebooks} are parsed to retain only each cell's type and source (i.e., code input), which removes the large spurious diffs in cell outputs produced by simply re-running a notebook.
  \item Binary artifacts, model checkpoints, logs, and files under the output directories carry only metadata, including path, change status, size, and the reason the contents were withheld.
\end{itemize}

\paragraph{Batching.}
Commits are grouped into units---one per commit, subdivided when a single commit's evidence exceeds a character budget---and packed greedily into requests until either the model's usable input context or the estimated output budget for the batch would be exceeded. In our descriptions, we abuse the word `commit' for `unit'.

\paragraph{Commit assessment example.} Suppose one commit introduces a new loss function, adds an independent evaluation comparison, and reformats a utility file. The intended output contains content in Table ~\ref{tab:commit_assessment_exapmle}.
\begin{table}[h!]
\centering
\small
\begin{tabular}{@{}ll@{}}
\toprule
Output & What it records \\
\midrule
Commit assessment  & \texttt{multiple}, referencing $a_1$ and $a_2$ \\
Decision $a_1$  & The loss-function choice, citing its relevant code \\
Decision $a_2$  & The evaluation choice, citing its relevant code \\
\bottomrule
\end{tabular}
\caption{Example of mapping a commit to decisions, where we suppose one commit introduces a new loss function, adds an independent evaluation comparison, and reformats a utility file.}
\label{tab:commit_assessment_exapmle}
\end{table}

\begin{table}[h!]
\centering
\small
\begin{tabular}{@{}llrr@{}}
\toprule
Assessment & Meaning & Count & Percentage \\
\midrule
\texttt{irrelevant} & no research decision         & 66,712 & 56.6\% \\
\texttt{support}    & evidence for an existing one & 17,640 & 15.0\% \\
\texttt{multiple}   & several decisions at once    & 16,544 & 14.0\% \\
\texttt{revise}     & changes an existing one      & 8,130  & 6.9\% \\
\texttt{new}        & begins a new one              & 5,130  & 4.4\% \\
\texttt{uncertain}  & held as an open thread       & 3,696  & 3.1\% \\
\midrule
\textbf{Total}      &                              & \textbf{117,852} & \\
\bottomrule
\end{tabular}
\caption{Commit assessments for 599 annotated pairs for NeurIPS 2022--2025.}
\label{tab:assessments}
\end{table}

\paragraph{Cost and throughput.} Regarding frontier LLM usage, the cost is input-dominated: input tokens to output tokens has a ratio 16:1. Table~\ref{tab:cost} shows that 99\% of LLM cost is in \texttt{commit\_scan}, the only step that transmits repository evidence.
\begin{table}[h!]
\centering
\small
\begin{tabular}{@{}lrrrr@{}}
\toprule
Function & Calls & Input Token / Call & Output Token / Call & Cost Percentage \\
\midrule
\texttt{commit\_scan}     & 8,543 & 143,042 & 9,175  & 99.0\% \\
\texttt{commit\_finalize} & 708   & 18,918  & 1,353  & 0.5\% \\
\texttt{annotation}     & 671   & 16,409  & 528    & 0.4\% \\
\midrule
\textbf{Total}          & \textbf{9,922} & & & \textbf{100\%} \\
\bottomrule
\end{tabular}
\caption{LLM cost by stage for NeurIPS 2022--2025.}
\label{tab:cost}
\end{table}

\begin{tcolorbox}[title=Prompt for annotation, colback=white, colframe=myblue, fonttitle=\bfseries]
\begin{VerbatimWrap}
You are analyzing a repository chronologically to recover research decisions observable in its public history.

PAPER CONTEXT
The paper title and abstract are supplied only to determine whether a repository change concerns the linked paper. They are not evidence for a research decision.
Do not reconstruct decisions, motivations, results, or methods from the paper.

RESEARCH DECISION
A research decision is a repository change that meaningfully affects the paper's proposed contribution, how that contribution is evaluated, or the settings and components of the proposed method.

Routine formatting, documentation, CI, dependency updates, bug fixes, and refactoring are not research decisions unless repository evidence shows that they alter the scientific method or evaluation.

CATEGORY
Give every decision exactly one category: `method`, `experiment`, or `ablation`. Label what the decision changes, not why it was made; wording such as "Hypothesize", "Investigate", or "Ablate" does not decide the category. Use the paper title and abstract only to identify the paper's contribution: what the paper builds or proposes (a model, objective, algorithm, estimator, theory, training or inference procedure, or system), or, for a benchmark, dataset, simulator, or platform paper, that released artifact. For an empirical-study or analysis paper, the contribution is the analysis, and the models it trains are experimental setup.
- `method`: the decision builds or revises the contribution itself.
  - This covers its architecture, representation, objective or loss form, training or inference procedure, estimator, data pipeline, theory, and claims about its scope.
  - For a system, it covers the system's operators and internals.
  - For a benchmark, it covers the benchmark's data, tasks, ground-truth targets, splits, environments, and construction rules.
  - Introducing a mechanism with its initial settings, replacing one mechanism with a different one, and creating a knob that is the paper's key novelty are all method.
- `experiment`: the decision changes how the contribution is evaluated or run for evaluation.
  - This covers evaluation datasets and conditions (data size, noise, dimension, domain), metrics and scoring, protocols, checkpoint selection, seeds, budgets, and reporting.
  - It covers baselines and comparators, including their hyperparameters. In a benchmark paper, the models run on the benchmark are baselines.
  - It covers training settings of the models an analysis paper studies, analyses and diagnostics, and comparisons against baselines.
- `ablation`: the decision changes the value of a hyperparameter of the proposed method, when that hyperparameter is not itself the paper's key novelty.
  - Hyperparameters include the training settings of the authors' own model: learning rate, batch size, epochs, optimizer or schedule, dropout, EMA, loss weights, temperatures, thresholds, sizes, and counts.
  - This holds whether the value is changed once, edited back and forth, or swept, and whether or not side-by-side runs are visible.
  - A controlled comparison among variants of the proposed method's own components (A vs B, with vs without) is also ablation.
\end{VerbatimWrap}
\end{tcolorbox}

\begin{tcolorbox}[title=Prompt for annotation (cont.), colback=white, colframe=myblue, fonttitle=\bfseries]
\begin{VerbatimWrap}
- Key-novelty knobs. This rule applies when the hyperparameter is itself the paper's key novelty. It also applies to a weight, strength, or level that controls the proposed term or mechanism, such that some value of it turns the method back into an existing baseline (for example, a weight of 0 that removes the proposed term). Creating the knob is `method`. A later change or sweep of its value is:
  - `experiment` if any value tried recovers an existing baseline, including when that value is one of several swept values;
  - `ablation` if no value tried does;
  - `experiment` if the evidence does not show whether any value recovers a baseline.
- Proposed metrics. When the contribution is itself a metric, score, evaluation method, or verifier, building it is `method`. Adopting or configuring standard metrics is `experiment`.
- Analysis papers. The paper's studied factor is `experiment`. A side-by-side comparison of the training recipe of the models it studies is `ablation`.
- Tie-break. A decision that both builds a mechanism and evaluates it takes the category of its main content:
  - `method` when that is building or revising the mechanism;
  - `ablation` when it is a controlled comparison of the method's own variants;
  - `experiment` when it is a comparison against baselines or an evaluation setting.

PROCESS
1. Process the supplied commits strictly in sequence order. Assess every commit. Do not silently omit one.
2. A commit may:
    - be research-irrelevant;
    - support an existing decision;
    - revise an existing decision;
    - begin a new decision;
    - combine several earlier commits into one decision; or
    - contain several distinct decisions when separate changes support them.
3. Merge engineering commits implementing the same scientific choice into one decision. A merge is transitive: if C was merged with A and later D supports C, D supports the existing {A,C} decision unless new evidence explicitly shows that the earlier grouping was wrong. Do not silently discard A or an intervening candidate B merely because C or D matched another thread.
4. Preserve uncertain or incomplete changes exclusively as evidence-backed entries in `open_threads` so later commits can clarify them. Never put an uncertain candidate in `decisions` with an open status. For example, if an engineering implementation that itself is not a decision but it could be a decision if later commits implement the remaining, that engineering implementation can be an open thread. Every new or updated thread must include the relevant commit SHAs and exact repository citations from the current call; the harness retains its earlier evidence locally. Uncertainty means the evidence is not yet sufficient to classify the scientific choice, not that the thread lacks evidence.
5. Never invent intent, causality, failed experiments, or observations.
6. Cite only supplied repository evidence. A diff excerpt must be copied exactly.
7. Preserve existing decision IDs. When merging decisions, record which IDs were merged rather than silently deleting them.
   `merge_log` may name only formal decision IDs supplied in `prior_state`. If several changes in the current call belong to one atomic decision, return that one decision directly; do not invent and immediately absorb temporary decision IDs.
8. A prior open thread may be resolved in three ways:
    - If it belongs to an existing prior decision, merge its commit SHAs and evidence into that decision, remove the thread, and record `merged_into_decision` with the existing decision ID.
    - If it becomes a distinct supported decision, remove the thread and record `promoted_to_decision` with its decision ID.
    - If later evidence shows that it is not a research decision, remove the thread and record `rejected` with an empty result decision ID.
9. The response is a delta, so omit unchanged prior open threads from `open_thread_upserts`; the harness retains them locally. Emit exactly one `thread_resolutions` entry only when the current call resolves and removes a prior thread. `thread_resolutions` contains only resolutions made in the current call; it is not a cumulative history.

DECISION VALIDITY AND OUTCOME
`status` records whether a candidate is genuinely a research decision. `outcome` records what later happened to a valid decision. Do not use one as a substitute for the other.
- Use `status: "supported"` for every genuine historical research decision, including choices that were later removed, disabled, or replaced.
- Use `status: "rejected"` only to remove an existing decision candidate that later evidence or final consolidation shows was never a research decision, such as setup-only work or an implementation step belonging to a parent choice. A rejected candidate must have `outcome: null`. Do not create a rejected decision for a newly observed irrelevant change, and use `merge_log` rather than rejection for duplicates.
- Use `outcome: "retained"` when the decision remains operative at the end of the observed history or no supplied evidence shows that it was abandoned or superseded.
- Use `outcome: "superseded"` only when a later extracted, supported decision clearly performs the same research role and repository evidence shows that it replaces the earlier choice. Preserve both decisions and set `superseded_by` to the successor's decision ID. A related later experiment, implementation change, deletion, or inactive invocation is not enough by itself.
- Use `outcome: "abandoned"` for every other discontinued decision, including one that is removed or disabled without such a clearly identified successor. When uncertain whether a later decision is a true replacement, use `abandoned`.
- Set `superseded_by` to null for every outcome other than `superseded`, and for every rejected candidate. Name only a decision ID that exists in this response or in `prior_state`; never invent one.
- A deletion, disabled invocation, or replacement can update a supported decision's outcome and add lifecycle evidence. It must not by itself change that decision's status to rejected.
- When a merge combines duplicate decision records, set the result's outcome from the latest lifecycle evidence. Upsert the merge target when its outcome changes.
\end{VerbatimWrap}
\end{tcolorbox}

\begin{tcolorbox}[title=Prompt for annotation (cont.), colback=white, colframe=myblue, fonttitle=\bfseries]
\begin{VerbatimWrap}
RECENT-FIRST MATCHING
For each new research-relevant commit, compare it first with open threads in `recent_first_matching_order.open_thread_ids`, then with established decisions in `recent_first_matching_order.decision_ids`. Stop at the first clearly compatible scientific choice. Recency is a search preference, not a requirement: a nonadjacent commit may join an older thread or decision when the semantic evidence is clear. Intervening unmatched threads remain open and independent. Reject a thread only when evidence shows that it is not a research decision or it remains unsupported during finalization. If evidence instead establishes that the thread was a genuine decision later abandoned or superseded, promote it with that outcome and, when superseded, with its successor's decision ID. Never reject it merely because another thread received a match.

ATOMIC DECISION GRANULARITY
Each decision must express exactly one independently meaningful scientific choice, comparison, or revision. Apply these tests consistently:
- Merge code changes, configuration edits, loaders, scripts, logging, and reporting utilities when they only implement or support the same method or experimental comparison.
- Keep two choices separate only when each independently changes the proposed method, a dataset comparison, the evaluation protocol, or an ablation. In this case, the successor formulation counts as an independent decision.
- A change in mathematical form, objective, architecture, training mechanism, inference procedure, or scientific data flow is a new decision even when it replaces—and cannot coexist with—the prior decision.
- A sweep over a single parameter, distribution, schedule, or component is one decision, however many commits set individual values; its category follows the CATEGORY rules. Record the swept range in the decision text rather than splitting it.
- The same rule applies across components: applying one baseline or normalization scheme to a second model, branch, or stage is support for the existing decision unless it introduces an independent research comparison of its own, e.g., augmenting the experiment evaluation protocol.
- A new dataset or model is a separate decision only when repository evidence shows that it introduces a distinct research comparison, not merely support code.
- A fallback, bug fix, parameter-counting utility, or result parser is not a separate decision unless it intentionally changes the scientific data, metric, protocol, or interpretation.
- Use one consistent level of scope within a benchmark. Do not combine many attack or model families into one decision while treating a comparable single integration as a separate decision without a scientific reason.
- Do not inflate the decision count by turning hyperparameters or implementation steps of exactly one scientific choice into separate decisions.

If a current decision becomes too broad, keep its independently variable choices separate while their full evidence is available. During finalization, merge over-split decisions and reject setup-only entries using the same tests. Preserve valid historical decisions with abandoned or superseded outcomes. Record every merge in `merge_log`.

INPUT
- paper_context: title and abstract
- prior_state: previously accumulated decisions and open threads
- recent_first_matching_order: open-thread and decision IDs ordered from most recently evidenced to oldest
- commits: the next chronological commit batch
- evidence records: selected nonempty textual evidence only, with evidence ID, SHA, date, path, change status, evidence kind, and retained content. A creation snapshot is the bounded file text from its first add event; later records for that path contain only diffs. Empty and manifest-only records remain in local evidence.json and are not supplied.

RETURN JSON
The response is a delta, not the complete rolling state.
- `decision_upserts` contains only decisions created, supported, semantically revised, assigned a new outcome, or explicitly rejected as invalid in this call. For an existing decision, return its complete updated semantic fields but only commit SHAs and exact citations newly supplied in this call. Omit unchanged prior decisions.
- `open_thread_upserts` contains only threads created or updated in this call, with only newly supplied commit SHAs and exact citations. Omit unchanged prior threads.
- `thread_resolutions` and `merge_log` contain only operations performed in this call.
- A `thread_resolutions` entry may name only an ID listed in `resolvable_open_thread_ids`. When that list is empty, return an empty `thread_resolutions` array.
- The harness retains and unions all omitted prior decisions, threads, commit SHAs, and citations locally.
\end{VerbatimWrap}
\end{tcolorbox}

\begin{tcolorbox}[title=Prompt for annotation (cont.), colback=white, colframe=myblue, fonttitle=\bfseries]
\begin{VerbatimWrap}
{
"commit_assessments": [
    {
    "commit_sha": "...",
    "classification": "irrelevant|support|revise|new|multiple|uncertain",
    "decision_ids": ["Dxxx"],
    "reason": "..."
    }
],
"decision_upserts": [
    {
    "decision_id": "Dxxx",
    "decision": "...",
    "category": "method|experiment|ablation",
    "status": "supported|rejected",
    "outcome": "retained|abandoned|superseded or null when rejected",
    "superseded_by": "decision ID of the successor, or null unless outcome is superseded",
    "commit_shas": ["..."],
    "evidence": [
        {
        "commit_sha": "...",
        "date": "...",
        "path": "...",
        "diff_excerpt": "..."
        }
    ],
    "why_research_relevant": "..."
    }
],
"open_thread_upserts": [
    {
    "thread_id": "Txxx",
    "summary": "...",
    "commit_shas": ["..."],
    "evidence": [
        {
        "commit_sha": "...",
        "date": "...",
        "path": "...",
        "diff_excerpt": "..."
        }
    ],
    "needed_evidence": "..."
    }
],
"thread_resolutions": [
    {
    "thread_id": "Txxx",
    "resolution": "merged_into_decision|promoted_to_decision|rejected",
    "result_decision_id": "Dxxx or empty when rejected",
    "reason": "..."
    }
],
"merge_log": [],
"batch_limitations": []
}

After the final batch, code verifies every SHA, path, and exact diff excerpt; consolidates duplicates; removes rejected candidates; retains supported decisions of every outcome; drops a `superseded_by` link whose successor did not survive validation; orders decisions by their first cited commit date; and derives decision_count itself. The annotation LLM then receives only those validated decisions and their evidence, not the entire accumulated conversation.
\end{VerbatimWrap}
\end{tcolorbox}

\begin{tcolorbox}[title=Judge rubrics, colback=white, colframe=purple, fonttitle=\bfseries]
\begin{VerbatimWrap}
You receive (1) an observable trajectory of research decisions, (2) the actual next decision, and (3) one prediction for the next decision. Evaluate how closely the prediction matches the actual decision. 

Score `component_match` (what part of the project is acted on), `operation_match` (the scientific action), and `specification_match` (concrete choices) independently from 0 to 2.

## Understand the decision before scoring

Describe each decision as an **operation on a component, with specifications**. Use the trajectory to resolve references and understand whether a choice is being introduced, extended, replaced, or tested. 

| Decision | Component | Operation | Specifications |
| --- | --- | --- | --- |
| Add CIFAR10 to the existing evaluation suite. | Evaluation dataset selection | Extend the suite | CIFAR10 |
| Replace the current loss with a weighted sum of classification and reconstruction losses. | Training objective | Replace the objective | The two loss terms and their weighted combination |
| Compare training with and without the reconstruction loss. | Training objective | Ablate a loss term | Reconstruction loss |

Identify the functional component; treat named implementations, such as AdamW versus SGD, as specifications.

## 1. `component_match`

Does the prediction act on the same functional part of the project as the actual decision?

**0 — A different part of the project, or no identifiable component.**

- One changes the optimizer, while the other changes training-data construction.
- One changes the model architecture, while the other changes the loss objective.

**1 — A related part, or only a broader identification of the correct part.**

- One changes an evaluation metric, while the other changes how uncertainty in that metric is estimated. Both concern measurement, but act on different parts of it.
- One changes token embeddings, while the other changes positional encodings. They concern related representation components.
- The actual decision changes the evaluation metric, but the prediction only says “improve evaluation,” without identifying a particular component.

**2 — The same functional component.**

- Both change the optimizer, even if they choose different optimizers.
- Both act on a hyperparameter search grid, even if one expands it and the other restricts it.

Tuning an optimizer and replacing it concern the same component. Identify experiments by the functional part they change, not their experiment numbers or dataset names.

## 2. `operation_match`

Does the prediction perform the same scientific action? Examples include introduce, replace, remove, extend, restrict, compare, ablate, calibrate, and validate. For questions and hypotheses, identify whether the statement asks a question, proposes an explanation, or tests it. 

**0 — A different or opposing operation, or no identifiable action.**

- One expands the hyperparameter search range, while the other narrows it.
- One adds a neural-network layer, while the other moves an existing layer to a different location.
- One proposes an explicit ablation, while the other merely says “improve the method” without identifying an action.

**1 — A related but distinct operation, or an underspecified version of the action.**

- One introduces a new tunable hyperparameter, while the other extends the search grid for an existing hyperparameter.
- One replaces the optimizer with a candidate optimizer, while the other benchmarks that candidate against the current optimizer before making a choice.
- The actual decision replaces a mechanism, while the prediction modifies that mechanism.

**2 — The same operation, including semantic paraphrases.**

- One introduces a baseline, while the other introduces a metric: both introduce, despite acting on different components.
- “Ablate the auxiliary loss” and “compare training with and without the auxiliary loss” both describe an ablation.

## 3. `specification_match`

Do the concrete choices agree, either exactly or at the functional-family level allowed by this rubric? Specifications include datasets, model families, architectures, loss terms, numeric thresholds, split ratios, and hyperparameter values. This dimension intentionally allows broader matches than exact names alone.
\end{VerbatimWrap}
\end{tcolorbox}

\begin{tcolorbox}[title=Judge rubrics (cont.), colback=white, colframe=purple, fonttitle=\bfseries]
\begin{VerbatimWrap}
**0 — No matching or meaningfully comparable concrete settings.**

- CIFAR10 versus ReLU: an evaluation dataset versus an activation function.
- Setting batch size to 32 versus 128: the component and operation match, but the values do not.
- The actual decision names AdamW, while the prediction says only “replace the optimizer”: no replacement optimizer is specified.

**1 — Partial agreement, or related settings used for different tasks.**

- CIFAR10 versus COCO object detection: both involve image data, but the tasks differ.
- The learning rate matches, but batch size differs or is omitted.

**2 — The defining settings agree.**

- CIFAR10 versus CIFAR100 when used as image-classification evaluation datasets.
- Rotten Tomatoes versus SST-2 when used as sentiment-classification evaluation datasets.
- ResNet versus ViT when the choice is a generic image-feature backbone and the architecture's internal mechanism is not the subject of the decision.

Apply family equivalence to corresponding roles, not to any shared technical term. ResNet and ViT can match as generic backbones, but not when convolution versus self-attention is itself being investigated.

If the actual decision supplies no concrete specifications, assign specification score 0 without lowering the other scores.

## Worked comparisons across all three dimensions

| Actual decision | Predicted decision | Component | Operation | Specification |
| -- | -- | -- | -- | -- | 
| Replace the optimizer with AdamW. | Switch the training optimizer to AdamW. | 2 | 2 | 2 | 
| Replace the optimizer with AdamW. | Replace the optimizer. | 2 | 2 | 0 | 
| Replace the optimizer with AdamW. | Compare AdamW against the current optimizer before selecting one. | 2 | 1 | 2 | 
| Replace the optimizer with AdamW. | Replace the tokenizer with BPE. | 0 | 2 | 0 | 
| Ablate the auxiliary loss. | Permanently remove the auxiliary loss to simplify training. | 2 | 1 | 0 | 
| Set batch size to 32. | Set batch size to 128. | 2 | 2 | 0 | 
| Increase batch size to 128. | Decrease batch size to 32. | 2 | 0 | 0 | 
| Set batch size to 32 and learning rate to 0.001. | Set batch size to 128 and learning rate to 0.001. | 2 | 2 | 1 |
| Add CIFAR10 to the evaluation suite. | Add COCO object detection to the evaluation suite. | 2 | 2 | 1 |
| Introduce a new evaluation metric. | Introduce an uncertainty estimate for the existing metric. | 1 | 2 | 0 | 

## Scoring discipline and output

- Score the semantic content of the `decision` sentence. If predicted `category`, `object`, or `operation` fields are present, use them only to clarify ambiguity. They cannot override the sentence or supply an action that it does not state. Predictions with only `category` and `decision` can be scored normally.
- Do not penalize component or operation scores for missing or different specification. They can receive 2 even when specification match is 0. Conversely, matching concrete names does not establish matching components or operations. Identify each dimension separately before assigning its score.
- Judge meaning rather than writing quality. 
- If a prediction contains several independent alternatives, do not select whichever one matches best. Judge what it actually commits to in each dimension.
\end{VerbatimWrap}
\end{tcolorbox}

\begin{table}[h!]
\centering
\small
\begin{tabular}{llll}
\toprule
Actual (A) and predicted (P) decisions & C & O & S \\
\midrule
\rowcolor{mygrey!10}\shortstack[l]{A: Replace the optimizer with AdamW.\\P: Switch the training optimizer to AdamW.}  & 2 & 2  & 2 \\
\shortstack[l]{A: Replace the optimizer with AdamW.\\P: Replace the optimizer.}   & 2  & 2  & 0 \\
\rowcolor{mygrey!10}\shortstack[l]{A: Replace the optimizer with AdamW.\\P: Compare AdamW against the current optimizer before selecting one.}  & 2  & 1  & 2 \\
\shortstack[l]{A: Replace the optimizer with AdamW.\\P: Replace the tokenizer with BPE.} & 0 & 2 & 0 \\
\rowcolor{mygrey!10}\shortstack[l]{A: Ablate the auxiliary loss.\\P: Permanently remove the auxiliary loss to simplify training.} & 2 & 1 & 0 \\
\shortstack[l]{A: Set batch size to 32.\\P: Set batch size to 128.} & 2 & 2 & 0 \\
\rowcolor{mygrey!10}\shortstack[l]{A: Increase batch size to 128.\\P: Decrease batch size to 32.} & 2 & 0 & 0 \\
\shortstack[l]{A: Set batch size to 32 and learning rate to 0.001.\\P: Set batch size to 128 and learning rate to 0.001.} & 2 & 2 & 1 \\
\rowcolor{mygrey!10}\shortstack[l]{A: Add CIFAR10 to the evaluation suite.\\P: Add COCO object detection to the evaluation suite.} & 2 & 2 & 1 \\
\shortstack[l]{A: Introduce a new evaluation metric.\\P: Introduce an uncertainty estimate for the existing metric.} & 1 & 2 & 0 \\
\bottomrule
\end{tabular}
\caption{Examples of comparing research decisions across three dimensions. `C' for \textit{Component}, `O' for \textit{Operation}, and `S' for \textit{Specification}.}
\label{tab:rubric_example}
\end{table}

\begin{figure}[h!]
\centering
\includegraphics[width=0.9\textwidth]{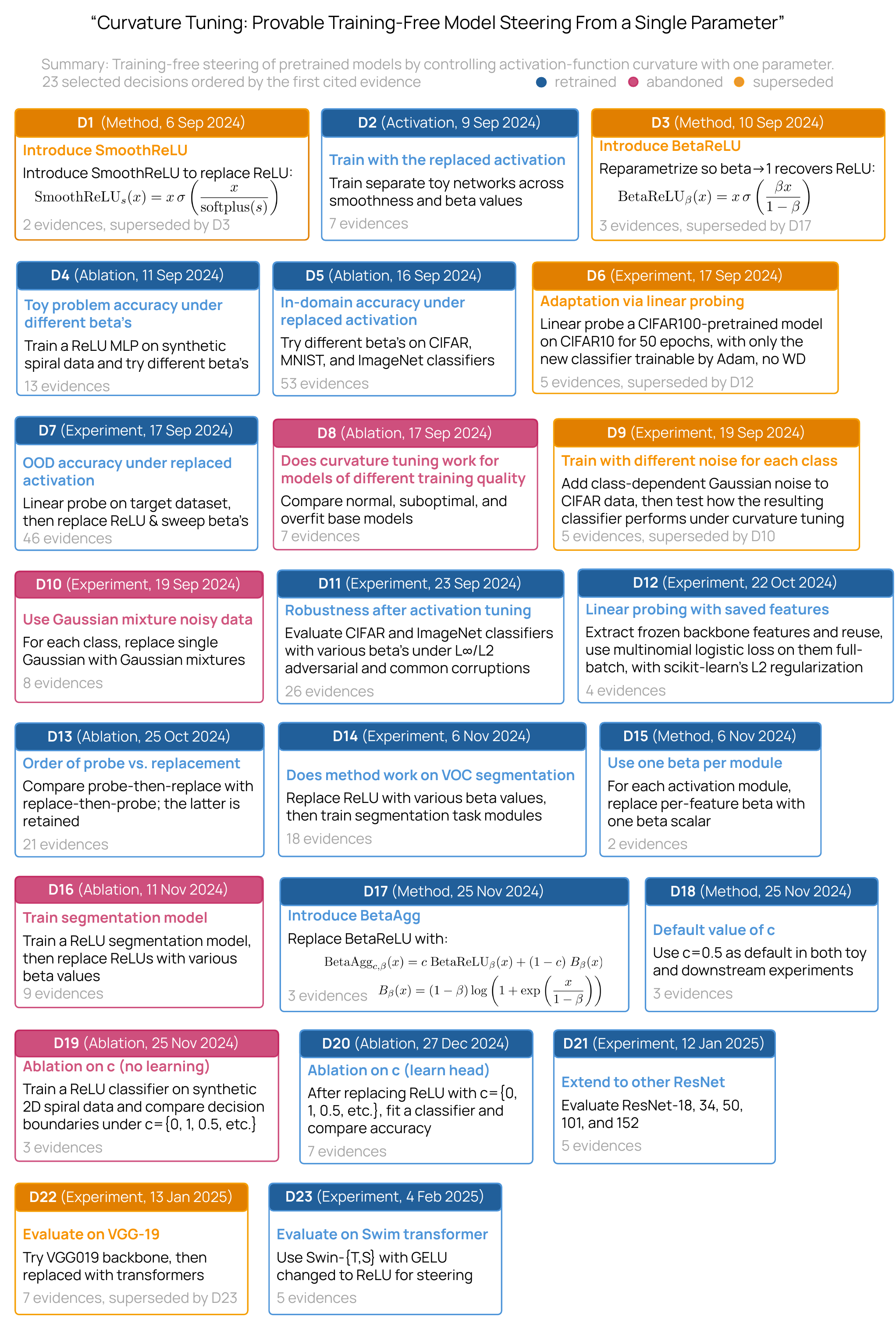}
\caption{Annotation example for \citet{hu2026curvature} (23 selected decisions). The annotation extracts methodological and experimental research decisions from the commit histories of its GitHub repository. Dates indicate the first repository evidence for each decision.}
\label{fig:example1}
\end{figure}

\begin{figure}[h!]
\centering
\includegraphics[width=\textwidth]{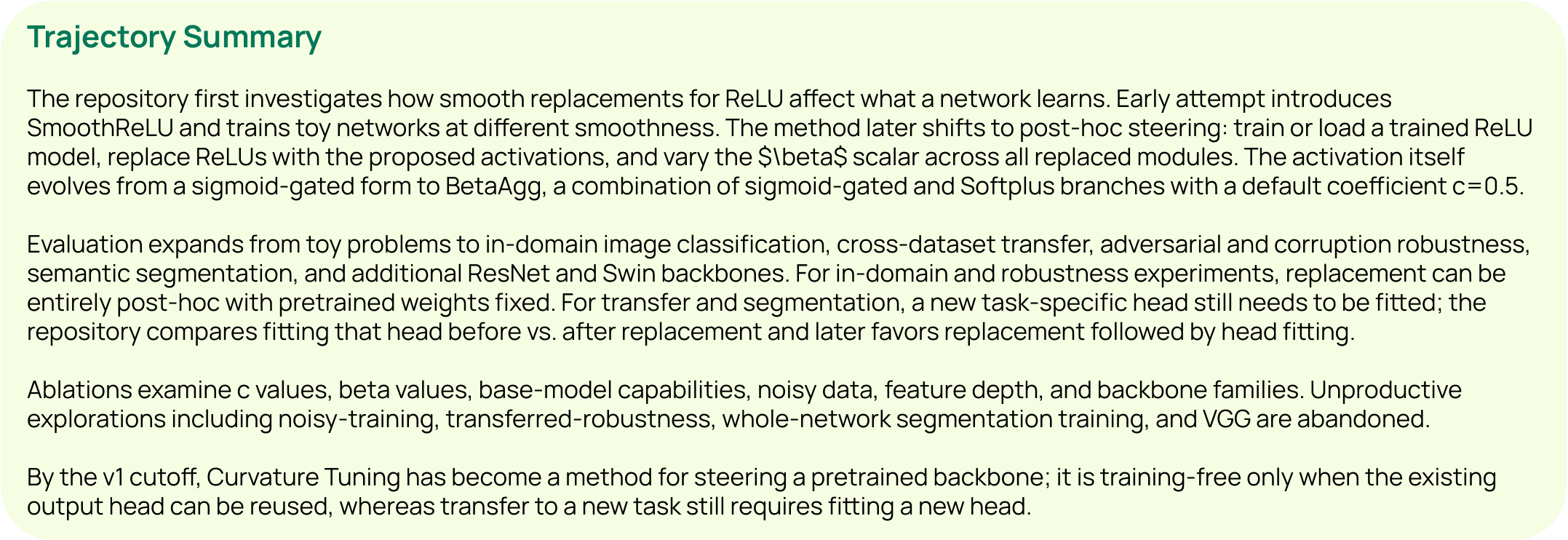}
\caption{Annotation example for \citet{hu2026curvature} (cont.).}
\label{fig:example2}
\end{figure}


\begin{figure}[h!]
\centering
\includegraphics[width=0.9\textwidth]{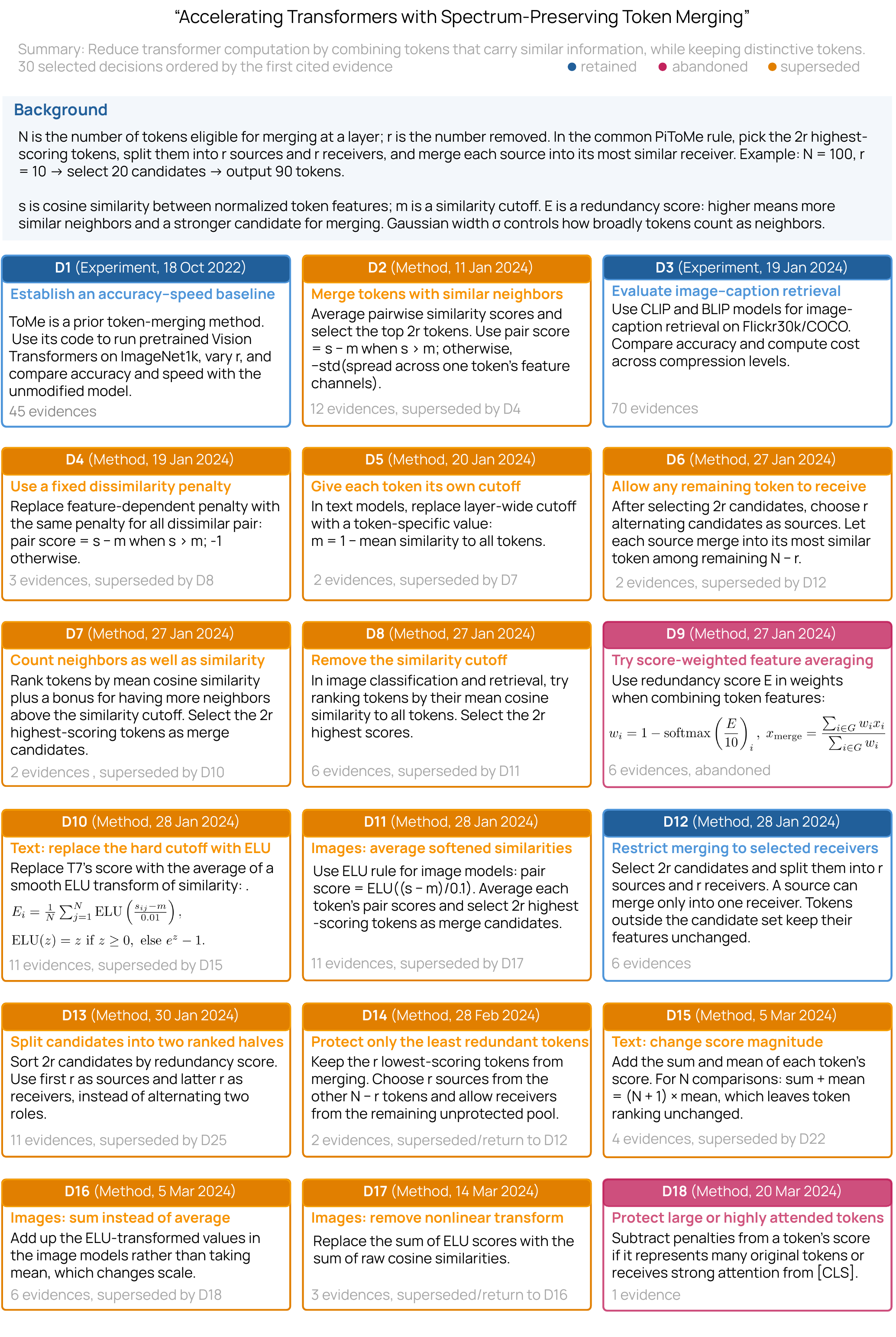}
\caption{Annotation example for \citet{tran2024accelerating} (30 selected decisions). The annotation extracts methodological and experimental research decisions from the commit histories of its GitHub repository. Dates indicate the first repository evidence for each decision.}
\label{fig:example4}
\end{figure}

\begin{figure}[h!]
\centering
\includegraphics[width=0.9\textwidth]{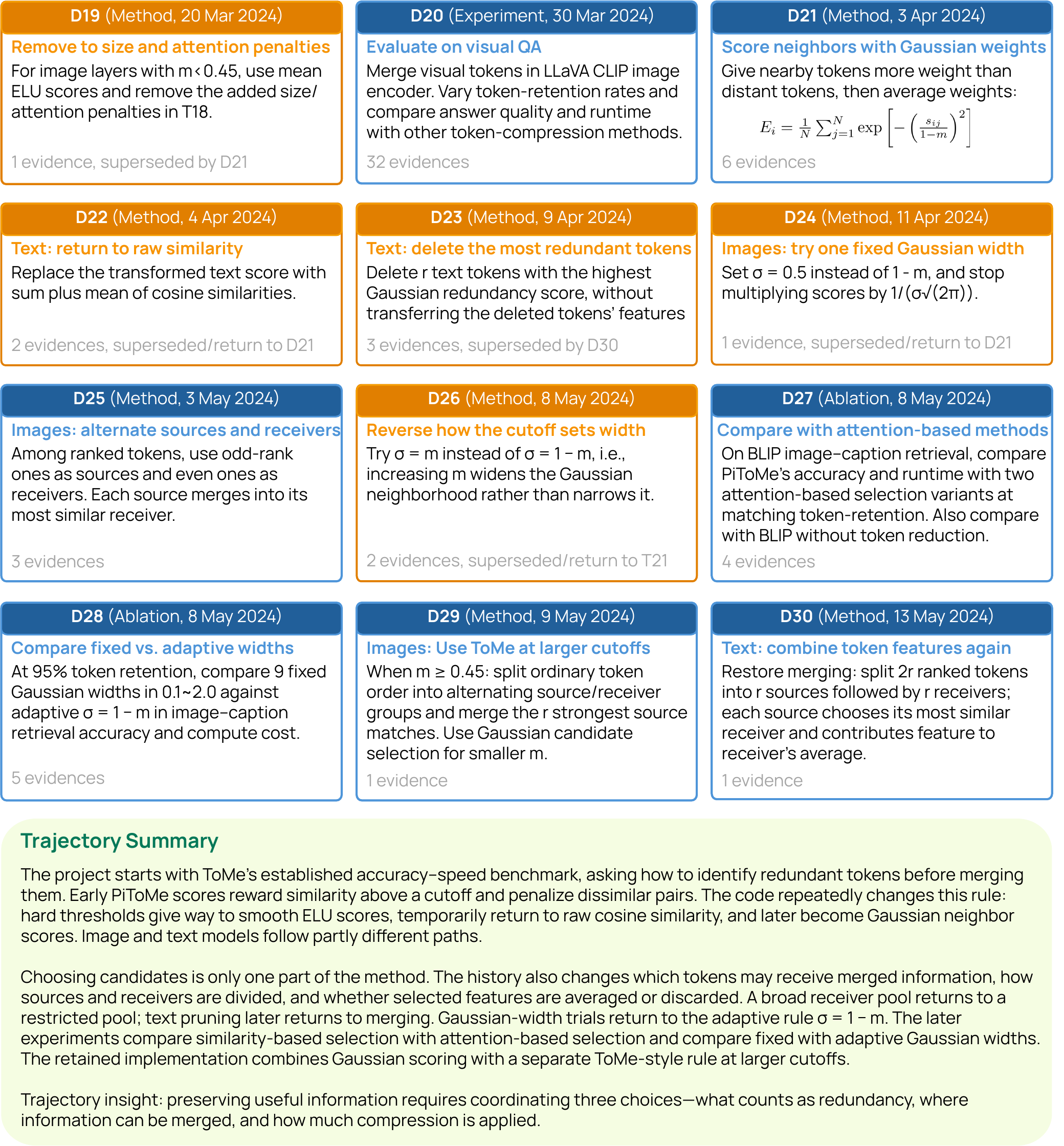}
\caption{Annotation example for \citet{tran2024accelerating} (cont.).}
\label{fig:example5}
\end{figure}

\begin{figure}[h!]
\centering
\includegraphics[width=\textwidth]{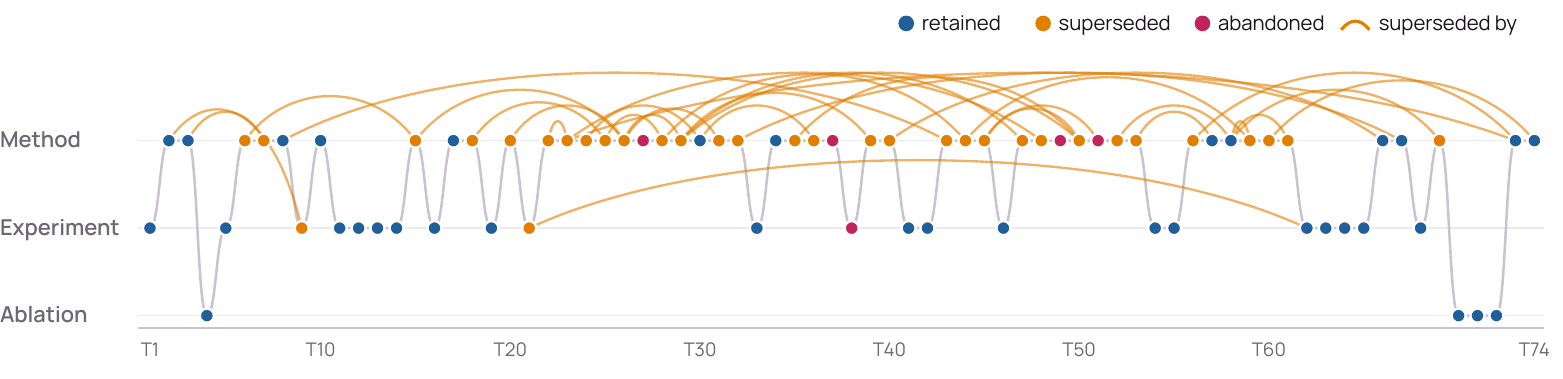}
\caption{Annotation example for \citet{tran2024accelerating}. It plots a trajectory by category, showing how a project moves between building its method, testing it, and ablating it. Orange arcs connect a superseded decision to the one that replaced it.}
\label{fig:example6}
\end{figure}
\section{Application Details} \label{sec:app:application}

\subsection{Test-Time Harness}
\paragraph{Judge variance and quality.}  
We experiment with judge GPT-5.6 Sol, observing that judge scores of fixed inputs have consistently small variance $<0.001$ (Table \ref{tab:judge_var}). Further, we evaluate judge quality by comparing judge scores with ground-truth scores: We synthesize paraphrases of the actual decisions, adjacent decisions to the actual ones, and unrelated decisions to the actual ones via varying \textit{Component}, \textit{Operation}, and \textit{Specification}. For use 100 samples from the evaluation set and generate paragraph, adjacent, and unrelated decisions for each sample using Opus 5 at high effort. During generation, Opus 5 receives the instructions as follows and does not see the judge rubric. 

\begin{tcolorbox}[title=Instructions for generating different types of predictions, colback=white, colframe=mygreen, fonttitle=\bfseries]
\begin{VerbatimWrap}
You create controlled candidate predictions for calibrating a research-decision judge. You receive an observable project trajectory and its actual next annotated decision.

Return four prediction objects:

1. `exact_actual`: Copy the actual decision sentence verbatim. Infer compatible category, component, and operation fields.
2. `faithful_paraphrase`: Express the same decision in genuinely different wording while preserving its component, scientific operation, and all defining concrete settings.
3. `adjacent`: Write a plausible decision in the same broad research area, but deliberately act on a different component and use a related but distinct scientific operation. It should merit component_match=1 and operation_match=1, not an exact match.
4. `unrelated`: Write a plausible decision that acts on a clearly different part of the project and performs a clearly different scientific operation. It should merit component_match=0 and operation_match=0.

Every decision sentence must agree with its category, component, and operation fields. Use only the allowed values in the schema. Do not explain the controls and do not mention their intended scores in their decision text.
\end{VerbatimWrap}
\end{tcolorbox}

We observe consistent matching between the obtained and expected scores (Table \ref{tab:judge_eval}).

\begin{table}[h!]
\centering
\small
\begin{tabular}{llcc}
\toprule
Judge Model & Metric & Sample SD & Fleiss' Kappa  \\
\midrule
\multirow{3}{*}{GPT-5.6 Sol} & \texttt{Score} & 0.032 & 0.719 \\
& \textit{Component} & 0.021 & 0.816 \\
& \textit{Operation} & 0.012 & 0.800 \\
\midrule
\multirow{3}{*}{Opus 5} & \texttt{Score} & 0.006 & 0.691 \\
 & \textit{Component} & 0.035 & 0.831 \\
 & \textit{Operation} & 0.040 & 0.778 \\
\bottomrule
\end{tabular}
\caption{Standard deviation and Fleiss' Kappa \citep{fleiss1971measuring} of the judge scores measured on fixed 100 baseline predictions.}
\label{tab:judge_var}
\end{table}

\begin{table}[h!]
\centering
\small
\begin{tabular}{llcccc}
\toprule
Judge Model & Predicted Decision & Intended \texttt{Score}  &  \texttt{Score} & \textit{Component}  & \textit{Operation}  \\
\midrule
\multirow{4}{*}{GPT-5.6 Sol} & Same & 4 & 4.00 & 2.00 & 2.00  \\
& Paraphrase  & 4 & 4.00 & 2.00 & 2.00  \\
& Adjacent  & 2 & 2.39 & 1.34 & 1.05  \\
& Unrelated & 0 & 0.10  & 0.04 &  0.06   \\
\midrule
\multirow{4}{*}{Opus 5} & Same & 4 & 4.00 & 2.00 & 2.00  \\
& Paraphrase  & 4 & 4.00 & 2.00 & 2.00  \\
& Adjacent  & 2 & 2.68 & 1.42 & 1.26  \\
& Unrelated & 0 & 0.12  & 0.04 &  0.08   \\
\bottomrule
\end{tabular}
\caption{Judge's scores match the intended scores when evaluating the actual decisions, their faithful paraphrases, adjacent decisions, and unrelated decisions.}
\label{tab:judge_eval}
\end{table}

\begin{table}[h!]
\centering
\small
\begin{tabular}{@{}llll@{}}
\toprule
Method  & Prefix-1 & Prefix-2  &  Prefix-3  \\
\midrule
Baseline  & 1.473 $\pm$ .088 & 1.555 $\pm$ .082 & 1.619 $\pm$ .049 \\
\textcolor{myyellow}{All demos}  &  1.537 $\pm$ .033 & 1.577 $\pm$ .049 & 1.602 $\pm$ .057  \\
\textcolor{mypink}{Random two demos} & 1.460 $\pm$ .062 & 1.505 $\pm$ .038 & 1.455 $\pm$ .110 \\
\textcolor{mygreen}{Retrieve two demos} & 1.590 $\pm$ .033 & 1.687 $\pm$ .068 & 1.714 $\pm$ .044  \\
\textcolor{myyellow}{Shuffled skill} & 1.575 $\pm$ .022 & 1.627 $\pm$ .071 & 1.555 $\pm$ .062 \\
\textcolor{myyellow}{Final paper skill} & 1.764 $\pm$ .030 & 1.490 $\pm$ .089 & 1.445 $\pm$ .024 \\
\textcolor{mygreen}{Skills} & \textbf{1.893} $\pm$ .037 & \textbf{1.781} $\pm$ .035  & \textbf{1.813} $\pm$ .042 \\
\bottomrule
\end{tabular}
\caption{Decision similarity of the predicted next decision based on previous $\{1,2,3\}$ decisions. Scores are judged by Opus 5 and averaged across evaluation trajectories, each with 3 runs. The conclusions from Table \ref{tab:harness} apply to the results in this table: Skills is best at every prefix, Retrieval beats Random at every prefix, and intermediate decisions provide learning signals beyound the final papers.}
\label{tab:harness_opus}
\end{table}

\begin{tcolorbox}[title=Skills, colback=white, colframe=myyellow, fonttitle=\bfseries]
\begin{VerbatimWrap}
# Early-Stage Next-Decision Prediction

Use this when the full observable history is one, two, or three decisions and the project is genuinely young. Nothing earlier is hidden; the prefix is the whole project so far. Your job is to name **one** next decision: its category (`method`, `experiment`, `ablation`), the **object** acted on, and the **operation**.

## Step 1 — Build the inventory from the prefix

Read the prefix and mark which of these exist:

- **Mechanism**: a proposed procedure, objective, loss, estimator, formulation, or model component.
- **Data**: a dataset, corpus, benchmark, or simulation/data-generation setup.
- **Measurement**: a metric, score, judge, diagnostic, or scoring protocol.
- **Tunables**: any parameter, threshold, temperature, radius, granularity, rate, count, or on/off switch explicitly named.
- **Revision signal**: a prefix item marked as superseded by a later observed item, or wording like "replace", "instead of", "reparameterize".

Almost every observed next decision acts on an object already present in the prefix or on the immediately adjacent missing slot of the same pipeline. Do **not** predict a new unrelated object, a new task domain, a theory/analysis section, or a baseline comparison with outside work unless the prefix already names it.

## Step 2 — Apply gap rules in priority order

1. **Data present, measurement absent → predict `experiment`: define how outcomes are scored.** This is the most reliable transition. Typical form: an evaluation protocol, error/quality metric, model-judge scoring, conditional-independence or correlation diagnostic, or a pass/fail criterion applied to the existing data. Also fires when the first decision is a mechanism plus a chosen corpus but no metric.

2. **Data and measurement both present, mechanism absent → predict `method`: introduce the project's core mechanism** (objective, estimator, training procedure) or the data-construction/labeling pipeline that feeds it. Prefixes that open with an evaluation suite, pilot comparison, or baseline comparison already containing metrics move to method next.
\end{VerbatimWrap}
\end{tcolorbox}

\begin{tcolorbox}[title=Skills (cont.), colback=white, colframe=myyellow, fonttitle=\bfseries]
\begin{VerbatimWrap}
3. **Mechanism present, no data and no measurement → choose between two continuations.** Predict another `method` decision when the prefix mechanism looks provisional or partial: a coarse prompt-based heuristic, a degenerate parameter trick, one stage of an obviously multi-stage system, or a component with adjacent uncovered stages (target construction, labeling, loss term, efficiency, resolution/granularity choice). The usual operation is *replace with a finer-grained variant of the same stage* or *add the next stage*. Predict an `experiment` setting up data/metrics instead when the mechanism reads as complete and self-contained (a full detector, formulation, or loss with nothing obviously missing).

4. **Mechanism just introduced with one salient continuous knob or discrete granularity, and no evaluation exists yet → predict `ablation`: sweep that knob.** Prefer this over rule 3 when the prefix foregrounds a parameter range, a smoothness/scale factor, or a resolution choice as the mechanism's defining degree of freedom.

5. **Mechanism plus measurement present, no ablation yet → predict `ablation` over the newest named tunable or the most recently introduced component.** Early ablations are always sweeps of, or on/off toggles of, something already written in the prefix (a threshold, a scale, a prompt variant, a sample population, a structural alternative), never novel machinery.

6. **Prefix ends on a narrow synthetic/toy experiment or ablation → predict `experiment` that widens scope**: more conditions, a broader or differently spaced parameter range, more datasets from the same family, or the same measurement transferred from the synthetic setting to a real pretrained/real-data setting.

7. **Revision signal present.** If a method stage has already been superseded and no evaluation of it exists, predict yet another `method` revision of that same stage. If an ablation or experiment already targeted the pre-revision version, predict re-running or extending that same ablation/experiment under the revised version.

8. **All-method prefixes of length two or three with no data or metric anywhere** continue in `method`, advancing along the pipeline (construction → granularity/scope → objective → estimation/efficiency/validation). Switch to `experiment` only when every core stage the prefix implies is already specified; then predict the dataset/condition configuration.

9. **All-experiment prefixes** continue in `experiment`, alternating between data/condition generation and diagnostics/metrics on those conditions, until both are covered.

## Step 3 — Fallback for ambiguous prefixes

If two rules conflict or none clearly fires, predict the **same category as the most recent decision, acting on the same object, as an incremental extension or replacement of it**. This is the modal behavior at all three prefix lengths. Prefer `method` over `experiment` when nothing in the prefix has yet been measured; prefer `experiment` when data exists but nothing has been scored.

## What cannot be inferred this early

State predictions at the object-and-operation level. Do **not** commit to:

- numeric sweep values, thresholds, learning rates, sample counts, seeds, or split boundaries;
- the identity of specific external models, tokenizers, or judge models;
- the full list of datasets in a later evaluation suite (name the family implied by the prefix instead);
- whether the next decision will be retained, superseded, or reverted;
- long-run architecture choices, theory results, or final comparisons.

When rule 5 or 6 fires, name the swept knob taken verbatim from the prefix vocabulary and say the range is underdetermined. When rule 1 fires, name the measurement type suggested by the prefix's object (e.g., ranking/threshold quality for detection-style objects, reconstruction/forecast error for generative or emulation objects, correctness scoring for question-answering objects) rather than inventing a specific metric formula.

## Output form

One sentence naming the category, then one or two sentences naming the object plus operation, then one clause stating the chief alternative you rejected and which inventory gap decided it. Keep specifications generic wherever the prefix does not pin them down.
\end{VerbatimWrap}
\end{tcolorbox}

\subsection{Two-Stage Training}

Denote $W_0$ as the pretrained model weights, and the weights after SFT are $W_{\mathrm{SFT}}=W_0+\Delta W_{\mathrm{SFT}}$. We initialize a policy $\pi_\theta$ by merging SFT LoRA into the $W_0$ model weights and then attaching a fresh LoRA adapter. Therefore, we have $W_\theta=W_{\mathrm{SFT}}+\frac{\alpha}{r}B_\theta A_\theta$. We freeze $W_{\mathrm{SFT}}$, initialize $A_\theta$ randomly and $B_\theta$ to zero, and optimize only the new LoRA weights. We freeze the reference policy  $\pi_{\mathrm{ref}}$ with weights $W_{\mathrm{SFT}}$.

Let $\mathcal{D}_{\mathrm{train}}=\{(h_i,a_i)\}_{i=1}^{N}$ contain research trajectory prefixes $h_i$ and their annotated next decisions $a_i$. Each prefix includes the fixed instruction and the preceding decisions. A generated decision $a$ is a sequence in JSON format with \texttt{category} and \texttt{content} fields. After SFT finishes, we use the resulting checkpoint as the frozen reference policy $\pi_{\mathrm{ref}}$. The reward is
\begin{equation}
  R_{ik}=
  \begin{cases}
    c_{ik}+o_{ik}, & \text{if $\hat a_{ik}$ is valid JSON and ends with EOS},\\
    0, & \text{otherwise}.
  \end{cases}
  \label{eq:grpo-reward}
\end{equation}
We retain a group only when $s_i>0$, which is computed by
\begin{equation}
  \overline{R}_i=\frac{1}{K}\sum_{k=1}^{K}R_{ik},\qquad
  \widehat{A}_{ik}=R_{ik}-\overline{R}_i,\qquad
  s_i=\sqrt{\frac{1}{K}\sum_{k=1}^{K}(R_{ik}-\overline{R}_i)^2}.
  \label{eq:grpo-advantages}
\end{equation}

Denote $p_{ikt}^{\Theta}=\pi_{\Theta}^\tau(y_{ik,t}| h_i,y_{ik,<t})$ as the probability of a sampled token under the current policy ($\Theta=\theta$), the policy that generated the batch ($\Theta=\mathrm{old}$), or the SFT reference ($\Theta=\mathrm{ref}$). 
We take a single update per batch, so GRPO's ratio clipping is inactive and omitted. With Dr. GRPO normalization \cite{liu2025understanding} and the GRPO reference penalty \cite{shao2024deepseekmath}, the loss is
\begin{equation}
  \mathcal{L}_{\mathrm{GRPO}}
  =\frac{1}{|\mathcal{S}|KC}
    \sum_{i\in\mathcal{S}}\sum_{k=1}^{K}\sum_{t=1}^{T_{ik}}
    \left[
      -\widehat{A}_{ik}\frac{p_{ikt}^{\theta}}{p_{ikt}^{\mathrm{old}}}
      +\beta\left(
        \frac{p_{ikt}^{\mathrm{ref}}}{p_{ikt}^{\theta}}
        -\log\frac{p_{ikt}^{\mathrm{ref}}}{p_{ikt}^{\theta}}-1
      \right)
    \right],
  \label{eq:grpo-loss}
\end{equation}
where $\mathcal{S}$ is the set of prefixes in the batch with non-constant rewards, $K$ is the group size, and $T_{ik}$ is the number of tokens in the generated decision $k$.

\begin{algorithm}[h!]
\caption{SFT Cold Start and Reinforcement Learning with Decision Similarity Rewards}
\label{alg:grpo}
\begin{algorithmic}[1]
\Require $\mathcal{D}_{\mathrm{train}}$, pretrained policy $\pi_\theta$, group size $K$
\Statex \texttt{// SFT Stage}
\State Train $\pi_{\theta}$ on $ \mathcal{D}_{\mathrm{train}}$ by minimizing $\mathcal{L}_{\mathrm{SFT}}$
\State Freeze copy $\pi_{\mathrm{ref}}\gets  \pi_{\theta}$
\Statex \texttt{// RL Stage}
\For{each batch $\mathcal{S}$}
  \State Sample $K$ decisions from $\pi_{\theta}^\tau(\cdot\mid h_i)$ for all $i\in \mathcal{S}$
  \State Compute rewards $R_{ik}$ via Eq.~(\ref{eq:grpo-reward}) and $\mathcal{L}_{\mathrm{GRPO}}$ via Eq.~(\ref{eq:grpo-loss})
  \State Update $\theta$ with $\nabla_\theta\mathcal{L}_{\mathrm{GRPO}}$
\EndFor
\State \Return $\pi_\theta$
\end{algorithmic}
\end{algorithm}

\paragraph{Implementation.}
We perform SFT for 5 epochs with batch size 16 and select the best checkpoint as the initialization for RL training. We perform RL updates for 10 epochs using batch size 8, $K=8$, $\beta=0.04$, $\epsilon=0.2$, $C=64$, and $\tau=0.8$. Our LoRA \citep{lora} uses rank $r=8$ and $\alpha=16$. We use a training dataset with 125 trajectories at with beginning prefix lengths \{1,2,3\}, yielding 375 training samples. We use the AdamW optimizer \cite{adamw} with a constant learning rate $10^{-4}$, no weight decay, and gradient norm clipping at $1$. We use GPT-5.6 Sol with medium reasoning as the judge.

\begin{table}[h!]
\centering
\small
\begin{tabular}{llcc}
\toprule
Model & GPT-5.6 Sol & Opus 5  \\
\midrule
Qwen3-8B base & 1.025 & 1.219 \\
SFT epoch 3 & 1.199 & 1.333 \\
GRPO epoch 7 & 1.505 & 1.674 \\
GRPO epoch 10 & 1.510 & 1.637 \\
\bottomrule
\end{tabular}
\caption{Decision Similarity Score of Qwen3-8B at different stages judged by GPT-5.6 Sol and Opus 5.}
\label{tab:training_cross_judge}
\end{table}

\end{document}